\documentclass{article} 

\usepackage{iclr2027_conference,times}

\usepackage{amsmath,amsfonts,bm}

\def\eqref#1{equation~\ref{#1}}

\def\1{\bm{1}}

\DeclareMathAlphabet{\mathsfit}{\encodingdefault}{\sfdefault}{m}{sl}
\SetMathAlphabet{\mathsfit}{bold}{\encodingdefault}{\sfdefault}{bx}{n}

\usepackage{amsmath,amssymb,amsthm}
\usepackage{graphicx}
\usepackage{booktabs}
\usepackage{multirow}
\usepackage{xcolor}
\usepackage{tikz}
\usepackage{enumitem}
\usepackage{tikz}
\usetikzlibrary{positioning,arrows.meta,fit,backgrounds,calc,shapes.geometric}
\usepackage{amsmath}
\usepackage{hyperref}
\usepackage{algorithm}
\usepackage{algorithmic}
\usepackage{hyperref}
\usepackage{algorithm}
\usepackage{hyperref}
\usepackage{hyperref}
\usepackage{url}
\usepackage{verbatim}
\usepackage{tikz}
\usetikzlibrary{shapes}
\usepackage{caption}
\usepackage{fix-cm}

\newcommand{\vir}{\textsc{ord-last}}
\newcommand{\virrq}{\textsc{ref-qspan}}
\newcommand{\virro}{\textsc{ref-out}}
\newcommand{\cers}{\textsc{CERS}}

\title{What Do Visual-Head Discovery Scores Actually Find? Attention Source, Necessity, and Random-Control Pitfalls in Vision-Language Models}

\author{
Md Abdul Kadir \& Daniel Sonntag \\
German Research Center for Artificial Intelligence (DFKI) \\
Saarbrücken, Germany \\
Oldenburg University \\
Oldenburg, Germany \\
\texttt{\{abdul.kadir, daniel.sonntag\}@dfki.de}
}

\iclrfinalcopy 
\begin{document}

\maketitle

\begin{abstract}
Attention heads that attend to image regions relevant to a question are increasingly used to
steer, ground and interpret vision-language models (VLMs) and to reduce their hallucinations. They
are mostly found by ranking heads with
an attention-pattern score and validated by steering or masking them. We ask how the choices
inside such a discovery score shape the heads it returns, and which of the two validations
identifies them. On Qwen3-VL-8B-Instruct we vary the query type, the discovery corpus and the
token positions whose attention is read (the \emph{attention source}) in a full $2\times3\times2$
design, and test all 12 head-sets for steerability (\cers: Contrastive Evidence-Region Steering score) and necessity (NCS: Necessity Contrast Score) against
layer-matched random head-sets. \textbf{(1)} The attention source and the query type reshape the
selected heads far more than the corpus: changing the source keeps a median of 49 of the top-100
heads, the query 55 and the corpus 84, while two halves of the same data share 96--99.
\textbf{(2)} Necessity follows the attention source: head-sets discovered from query-span
attention are more necessary than last-token ones (difference $0.083$, $p{=}6{\times}10^{-10}$),
and the two referring + query-span sets lie 17--19 standard deviations above 250 layer-matched
random sets (Holm-adjusted $p{=}0.048$). The referring + query-span effect replicates on several recent VLMs.
\textbf{(3)} Steerability is not specific to the
discovered heads: no discovered head-set steers
better than its layer-matched random sets (rank $p\ge0.14$), and random sets drawn from the same
or from any layer span the discovered sets' scores. Necessity therefore separates the discovery
settings, while steering at this strength does not. We turn
these findings into a protocol for reporting visual-head discoveries.
\end{abstract}

\section{Introduction}

\begin{figure}[t]
\centering
\definecolor{figApink}{RGB}{253,242,242}\definecolor{figAred}{RGB}{159,33,18}\definecolor{figInk}{RGB}{12,28,43}%
\setlength{\unitlength}{0.0013333\linewidth}%
\begin{picture}(750,520.5)
\put(0,0){\includegraphics[width=\linewidth]{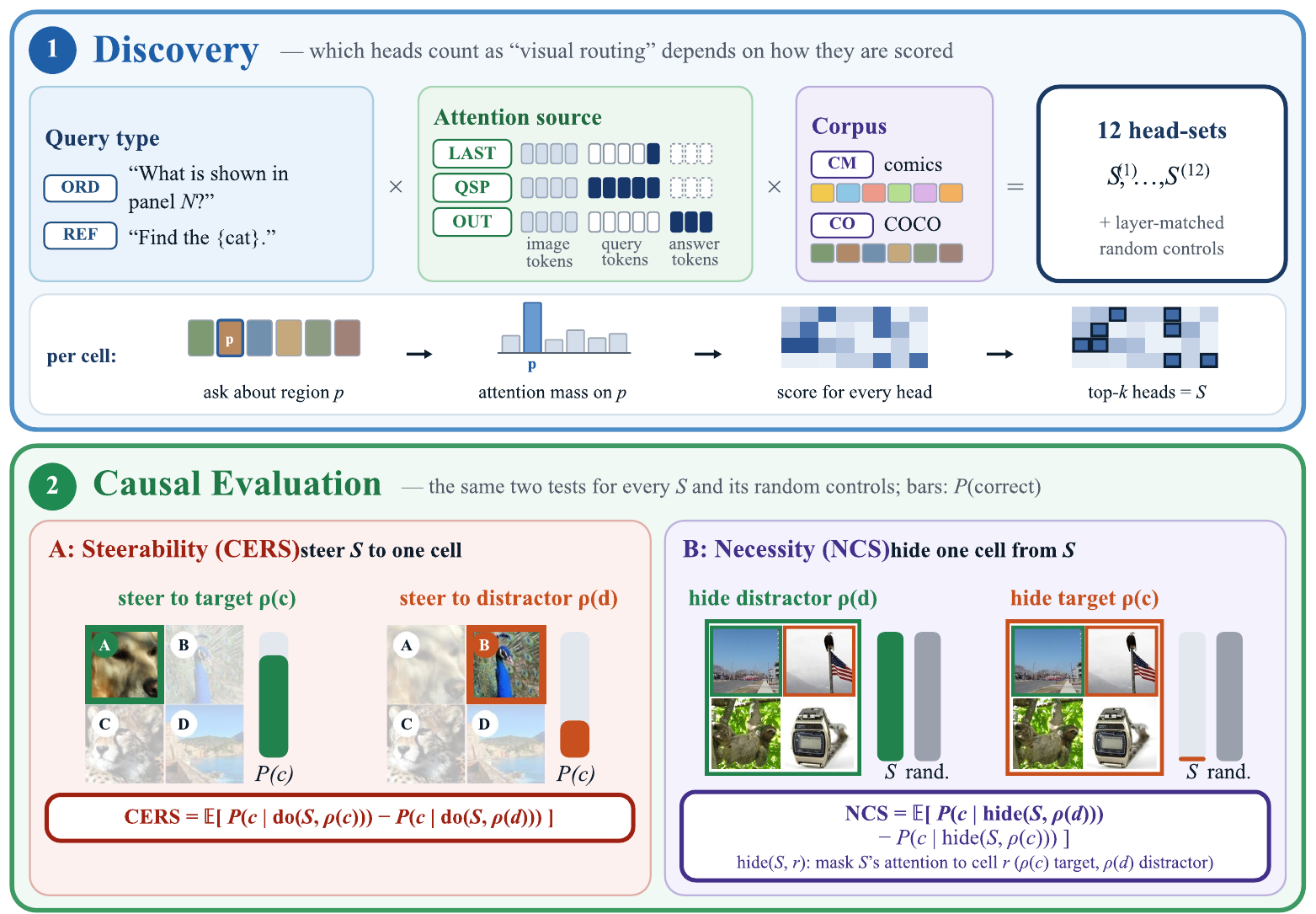}}
\put(46.5,191){\textcolor{figApink}{\rule{162\unitlength}{25\unitlength}}}
\put(48.4,206.6){\makebox(0,0)[l]{\textcolor{figAred}{\fontsize{9.4}{10}\selectfont\bfseries Steerability (CERS)}}}
\put(580,284){\textcolor{white}{\rule{108\unitlength}{21\unitlength}}}
\put(686.3,297.4){\makebox(0,0)[r]{\textcolor{figInk}{\fontsize{6.8}{7}\selectfont top-$k$}}}
\end{picture}
\caption{Overview. \textbf{(1) Discovery.} Two query types (ordinal and referring; prompt shortened in the figure) $\times$ three attention sources (last prompt token, query-span tokens, generated
answer tokens) $\times$ two corpora (comics, COCO) give 12 discovery scores. For each, we ask
about one region $p$ of a six-panel strip, record how much attention every head sends to
$p$, and rank the heads; the head-set $S$ is the top-100 heads, and each $S$ is paired with
layer-matched random controls. \textbf{(2) Causal evaluation.} Every $S$ and its controls get
the same two tests. \emph{Steerability} (\cers, Eq.~\ref{eq:cers}): steer $S$ onto the cell
holding the correct answer ($S{\to}c$) or a distractor's cell ($S{\to}d$) and compare
$P(c)$. \emph{Necessity} ($\mathrm{NCS}$, Eq.~\ref{eq:ncs}): mask $S$'s attention to all image tokens on a matched vs.\ mismatched ImageNet MCQ. The bars illustrate the related grid contrast $\mathrm{NCS}_{\mathrm{grid}}$ (Eq.~\ref{eq:ncs-grid}): hide a distractor's cell $\rho(d)$ or the answer's cell $\rho(c)$ from $S$; $P(c)$ for $S$ and a random head-set on one real item.
}
\label{fig:method-overview}
\end{figure}

A vision-language model (VLM) reads an image and a prompt together and writes an answer through
many attention heads \citep{aswani2017attention,alayrac2022flamingo}. A growing body of work finds
small sets of heads that attend to the image, or to the region a question refers to
\citep{bi2025unveiling,kang2025localization,kim2025interpreting,sun2026core,gazeheads2026,vrh2026},
and uses them to ground answers and reduce hallucination
\citep{kang2025localization,sandhu2026targeting}. Most of these heads are found by ranking heads
with an attention-pattern score. Where a head attends is a correlation, so the
evidence that the model \emph{uses} a head-set has to come from an intervention.

Two interventions answer two different questions. \emph{Steering} overrides, for one forward
pass, where a head-set attends (boosted on one image region, suppressed on the rest); it asks
whether the heads are a lever on the answer (\emph{steerability}). \emph{Masking} stops the
head-set from attending to the image at all; it asks whether the model needs the heads
(\emph{necessity}). The two are independent: a head-set can be a lever the model does not rely
on, or a route the model needs but that is too redundant to redirect on its own. We therefore
score every head-set with both, as \cers\  (Contrastive Evidence-Region Steering) and NCS (Necessity Contrast Score) (Figure~\ref{fig:method-overview}).

Validation varies: some studies characterize the heads by their attention patterns
\citep{bi2025unveiling}, some ablate top-ranked against lower-ranked heads \citep{sun2026core},
and some compare with random heads matched in layer \citep{sandhu2026targeting}. Two concurrent
2026 papers, one validated by steering and one by masking, show why such claims are hard to
compare. \emph{Gaze Heads} \citep{gazeheads2026} are found with an
attention-mass score on comic-panel tracking and validated by steering (answers follow the steered panel $83.1\%$ of the time); in the
released code, the free-text answers are scored by an LLM judge that must always pick a panel,
and the random baseline is a single draw of 100 heads from layers 20--35. \emph{Visual
retrieval heads} \citep{vrh2026} are found with a region-attention score on grounding data and
validated by masking alone, against a single draw of 20 random heads; a masking drop by itself cannot separate heads that carry question-relevant
evidence from heads that are generically important. Because the two papers differ in
intervention, task, metric and control, their claims cannot be compared, and neither varies
the design choices hidden inside its own discovery score.

We apply both interventions to every head-set that a full $2\times3\times2$ variation of the
discovery score produces, read a deterministic forced-choice probability (no generation, no
judge), contrast paired steers of the same heads on the same item, and rank every score
against many depth-matched random head-sets. The four main results are:
\begin{enumerate}
\item \textbf{The attention source and query type reshape which heads are found}
(\S\ref{sec:mechanism-main}). Of the top-100 heads, changing only the attention source keeps a
median of 49 (36--79), the query type 55 (35--79) and the corpus 84 (78--91), against 96--99
for two random halves of the same data.
\item \textbf{Necessity follows the attention source} (\S\ref{sec:ncs-main-result}).
Query-span head-sets are more necessary than last-token ones (NCS difference $0.083$,
$p{=}6{\times}10^{-10}$), no last-token head-set on Qwen3-VL-8B beats its random controls, and the two
referring + query-span sets lie 17--19 standard deviations above 250 layer-matched random
sets, while head-sets built to cause generic damage score at most $0.025$ against $0.125$.
\item \textbf{At the default strength, steerability is not specific to the discovered heads}
(\S\ref{sec:ncs-main-result}). No discovered head-set steers better than its layer-matched random sets (rank $p\ge0.14$);
random sets from the same or from any layer span the discovered sets' scores, and single
random draws range from $-0.34$ to $+0.84$ (layer-matched) and $-0.36$ to $+0.77$ (any layer), so the single-draw baselines used in prior work
cannot place a head-set within this distribution.
\item \textbf{A reporting protocol for visual-head claims} (\S\ref{sec:sanity-check},
\S\ref{sec:discussion}). Layer-matched random heads reach 73\% of a discovered set's
steering effect on the 11-task leak-audited subset of a 14-task pointing suite, so steering claims need many matched random
draws; necessity claims need controls that separate targeted from generic damage; and converted
benchmarks need leak checks before head-sets are compared.
\end{enumerate}


\vspace{-10pt}
\section{Related Work}
\label{sec:related}

\textbf{Functional heads.} Some heads perform specific functions: induction heads
\citep{olsson2022induction}, long-context retrieval heads \citep{wu2024retrieval}, and a few
critical heads among many prunable ones \citep{michel2019sixteen,voita2019analyzing}. In VLMs,
visual heads \citep{bi2025unveiling}, localization heads \citep{kang2025localization}, CoRe heads
\citep{sun2026core}, Gaze Heads \citep{gazeheads2026} and visual retrieval heads \citep{vrh2026}
are selected by where heads attend; head attribution \citep{kim2025interpreting} instead scores
each head's contribution to image-to-text information flow. Visual attention sinks
\citep{kang2025toldvisualattentionsink} attend heavily to image tokens without carrying task content, so high
image attention alone does not show use.

\textbf{Validating visual heads.} A clean causal test needs an intervention, a deterministic
readout, a contrast that cancels the side effects of the intervention, and random baselines
matched in depth and drawn many times. The two published validations each meet part of this:
steering read out by an LLM judge in the released code \citep{gazeheads2026} and masking
\citep{vrh2026}, each against a single random draw. Neither tests the other's heads, so
it is unknown whether ``gaze'' and ``retrieval'' heads are one mechanism. \cers\ and NCS apply
both tests to the same heads under the same readout and controls.

\textbf{Causal localization.} Existing methods intervene on internal representations to test
whether a component mediates a behavior. Causal tracing \citep{meng2022locating} and
activation or path patching
\citep{wang2023ioi,goldowskydill2023path,heimersheim2024activation,zhang2023best} do this by
replacing internal states across factual and counterfactual inputs, following the logic of
causal mediation \citep{vig2020causal}. Interchange interventions \citep{geiger2021causal}
and causal scrubbing \citep{chan2022scrubbing} instead test more structured hypotheses about
which information a component must carry. Attribution patching
\citep{syed2023attribution,kramar2024atp} and circuit discovery \citep{conmy2023acdc} scale
the same idea; MIB \citep{mueller2025mib} benchmarks it. Faithfulness still depends on the
ablation and the metric \citep{miller2024faithfulness,zhang2023best}, and a localized
component need not be the place an edit works \citep{hase2023localization}. Our tests reuse
these tools on image attention: masking is attention knockout \citep{geva2023dissecting}
restricted to image tokens, as in VLM information-flow studies
\citep{neo2024towards,zhang2025crossmodal,basu2024understanding}; the matched/mismatched
arm follows semantic image corruption \citep{golovanevsky2025notice}; steering is PASTA's
attention-bias edit \citep{zhang2024pasta}. The difference is the object of the test: we
score whole head-sets found by an attention-pattern score, against many layer-matched random
sets, on converted pointing items from CV-Bench \citep{tong2024cambrian}, MMStar
\citep{chen2024mmstar}, MathVista \citep{lu2024mathvista}, CountBenchQA
\citep{paiss2023teaching}, SpatialRGPT-Bench \citep{cheng2024spatialrgpt}, Spatial457
\citep{wang2025spatial457} and VAW \citep{pham2021vaw}.

\section{Discovery Scores and Causal Tests}
\label{sec:method}

\subsection{Discovery scores and the attention source}
\label{sec:discovery}

A discovery score ranks heads by how much attention they send to the image tokens of a
known target region $T$ while the model processes a query. For head $(\ell,h)$,
\begin{equation}
\label{eq:discovery}
M_{\ell,h} \;=\; \mathbb{E}_{x}\Big[\tfrac{1}{|Q_x|}\textstyle\sum_{i\in Q_x}\sum_{j\in T_x} A^{(\ell,h)}_{ij}(x)\Big],
\end{equation}
where $A^{(\ell,h)}(x)$ is the head's attention matrix on example $x$ and $Q_x$ is the set of
\emph{source} positions: the attending rows of $A$ that are averaged \cite{gazeheads2026,vrh2026}. The score bundles three choices. The
\textbf{query} is either \emph{ordinal} (``What is happening in the $N$-th panel from the
left?'', asked once for every panel of every strip) or \emph{referring} (``Find the
truck.'', asked once per strip). The \textbf{attention source} $Q_x$ is the last prompt token
(\textsc{last}), the query-span tokens (\textsc{qspan}), or the generated answer tokens
(\textsc{out}). The \textbf{corpus} supplies images and target regions; our corpora are
$1\times6$ horizontal strips of comic panels \cite{gazeheads2026} or COCO photos \cite{cocoimage} (``strips'') and they are highly unrelated. Gaze Heads use
ordinal + last-token, which we call \vir; our implementation matches an independent re-implementation
of their score exactly (100/100 top-100 overlap on the same corpus). The visual-retrieval-head score uses referring +
query-span (\virrq) or referring + output-token (\virro); \citet{vrh2026} report the
output-token variant as their best. The levels of each factor are taken from these two papers
rather than chosen freely: both query types and all three attention sources are their native
choices, the comics corpus is Gaze Heads' discovery data, and COCO photos with object boxes are
the grounding data of our visual-retrieval-head reproduction. Crossing them makes every
published setting one of the 12. Head-sets are the
top-$k$ heads of $M$, either flat over the whole matrix or with a fixed per-layer quota
$k_{\mathrm{per\_layer}}$ (Appendix~\ref{app:algorithm}); the size and rule used for every
table are listed in Appendix~\ref{app:protocol}. Figures use the axis labels
``ord-last'', ``ref-qspan'', and ``ref-out'' for \vir, \virrq, and \virro\ respectively;
Appendix~\ref{app:twelvecell-method} defines all 12 settings.

\subsection{Steerability: the \cers\ contrast}
\label{sec:steerability}

Each test image comprises $n{=}4$ images arranged in a $2\times2$ grid. Each answer option $o\in{A,B,C,D}$ corresponds to a region $\rho(o)$ with a known ground-truth label. In the pointing suite, letters identify the individual cells. This design increases task difficulty while providing clearer spatial separation between regions and reducing distributional interference \cite{kadir2026conceptguided}. We read the next-token
probabilities of the four letter tokens, renormalized to sum to one ($P(\cdot)$, the
\emph{forced probability}), from one forward pass; no text is generated and no judge is used.
Steering head-set $S$ toward region $r$, written $\mathrm{do}(S,r)$, adds a bias to the
pre-softmax attention logits of every head in $S$ at every attention row:
\begin{equation}
\label{eq:steer}
\tilde{z}^{(\ell,h)}_{ij} = z^{(\ell,h)}_{ij} + \beta\,\big(\mathbb{1}[j\in I_r] - \mathbb{1}[j\in I\setminus I_r]\big),\qquad (\ell,h)\in S,
\end{equation}
where $I$ is the set of image-token positions, $I_r$ those of the region $r$ (due to the grid design we can map the region to image tokens without an additional bounding box), and $\beta=10^4$;
this boost/suppress form and strength are the defaults of the released Gaze Heads code
\citep{gazeheads2026}, and masking uses the same $\beta$.
With correct letter $c$ and a random distractor $d\neq c$, the
\textbf{Contrastive Evidence-Region Steering score} (\cers) is
\begin{equation}
\label{eq:cers}
\mathrm{CERS}(S) \;=\; \mathbb{E}\Big[\, P\big(c \mid \mathrm{do}(S, \rho(c))\big) \;-\; P\big(c \mid \mathrm{do}(S, \rho(d))\big) \,\Big],
\end{equation}
written $P(c\mid S{\to}c)-P(c\mid S{\to}d)$ in Figure~\ref{fig:method-overview}
(worked example in Appendix~\ref{app:cers-ncs-diagram}). Both terms use
the same image, prompt and letter order and the same intervention, so the model's unsteered
preference for $c$ and any direction-independent side effect cancel. \cers\ measures
steerability: every other head still sees the whole image. Random heads can steer
too (\S\ref{sec:sanity-check}), so \cers\ is always read against random head-sets.

\subsection{Necessity: the NCS masking contrast}
\label{sec:necessity-method}

Masking $S$ adds $-\beta$ to its attention logits toward all image tokens. On a four-way ImageNet
multiple-choice question (MCQ; ``Which of the following is shown in this image?'', letter
order randomized) we apply the same mask to a \emph{matched} image (the target class is
visible) and a \emph{mismatched} image (an independently sampled photograph of a class
excluded from all candidate answers). The \textbf{Necessity Contrast Score} is the
difference of the two probability drops: matched drop minus mismatched drop.
Generic masking effects cancel in the contrast. On this MCQ the mismatched drop sits near
zero, so NCS is the matched-image drop up to noise (Appendix~\ref{app:d19}).
Figure~\ref{fig:method-overview} also shows a region contrast on $2\times2$ grid tasks,
where $\mathrm{hide}(S,r)$ masks the attention of $S$ to the tokens of cell $r$ only;
both terms mask an equal-sized cell, so generic effects of masking cancel within each item
(Appendix~\ref{app:clean-grid}):
\begin{equation}
\label{eq:ncs-grid}
\mathrm{NCS}_{\mathrm{grid}}(S)=\mathbb{E}\Big[P\big(c\mid\mathrm{hide}(S,\rho(d))\big)-P\big(c\mid\mathrm{hide}(S,\rho(c))\big)\Big].
\end{equation}
NCS therefore measures the matched-image drop on a probability scale; generic damage is controlled
with random head-sets and a stress test: attention-sink heads, text-heavy heads, whole layers and random
sets of 200 or 400 heads score NCS $-0.001$ to $0.025$ against $0.125$, and a variant whose
control arm is a text-solvable question gives $0.002$--$0.010$ against $0.112$, although masking
those sets' text attention lowers that arm by $0.23$--$0.78$ (Appendix~\ref{app:e2-stress}).

\section{Experimental Setup}
\label{sec:setup}

\textbf{Models.} Qwen3-VL-8B-Instruct for the 12-setting design and every main-text table;
second-model checks on Qwen3-VL-4B and Qwen2-VL-7B; further models in
Appendix~\ref{app:crossmodel}, \ref{app:d11}. \textbf{Discovery.} Eq.~\ref{eq:discovery} on
$N{=}400$ strips per corpus; head-set sizes (top-20, top-100, or
$k_{\mathrm{per\_layer}}\in\{1,2,3\}$) are stated per table and listed with every other
setting in Appendix~\ref{app:protocol}. \textbf{Controls.} Layer-matched random head-sets
copy $S$'s per-layer counts and exclude its heads: $R{=}20$ per setting, $R{=}250$ for the two
referring + query-span settings, one draw on the pointing suite and the out-of-domain benchmarks. Rank $p=(b+1)/(R+1)$, with $b$
the number of controls scoring at least as high as $S$ (floor $0.048$ at $R{=}20$, $0.004$ at
$R{=}250$). Random head-sets additionally matched on each head's
image-attention mass remove none of the necessity effects (all 12 settings,
Appendix~\ref{app:twelvecell-e1}). \textbf{Statistics.} $95\%$ $t$-intervals over the 400 items for one fixed head-set;
Holm correction over the 12 settings; factor-level and shared-vs.-specific tests use the same
400 items as the unit.

\section{Results}
\label{sec:results}

\subsection{Discovery choices reshape which heads are found}
\label{sec:mechanism-main}

\begin{figure}[t]
\centering
\includegraphics{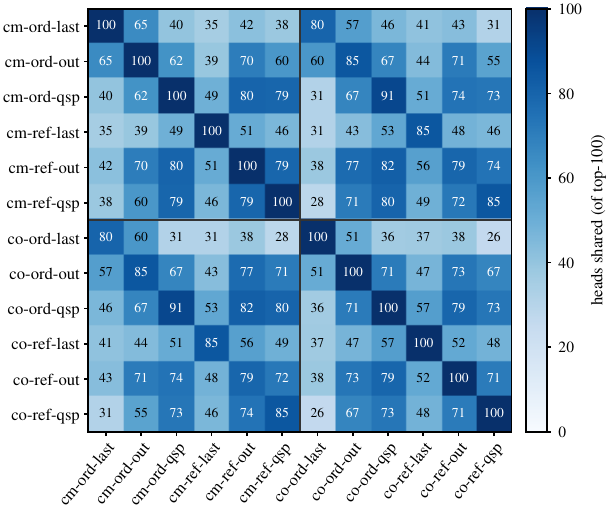}
\vspace{-2mm}
\caption{Top-100 head overlap between all 12 discovery settings, Qwen3-VL-8B. The same
query and attention source agree across corpora (e.g.\ cm-/co-ref-qsp share 85/100), while
changing the attention source moves most of the set (cm-ord-last vs.\ cm-ref-qsp: 38/100;
minimum 26/100).}
\label{fig:overlap-12cell}
\end{figure}

Two random halves of the same discovery data share 96--99 of the top-100 heads, but changing
the corpus keeps a median of 84 (78--91), the query type 55 (35--79) and the attention source
49 (36--79) at the same sample size (Table~\ref{tab:noise-main}, Figure~\ref{fig:overlap-12cell}). To isolate the attention
source we hold corpus and layout fixed (200 COCO strips) and cross query type with source
(Table~\ref{tab:crossover-main}). Gaze Heads' own query read from the query span instead of
the last token overlaps the retrieval-style native sets ($10/20$, $14/20$) more than native
\vir\ ($7/20$); switching the query with the source fixed at the last token drops overlap with
native \vir\ from 19 to 5. The two levers are of similar size, and the corpus matters least.
The published families are near-disjoint at top-20 on this model (2/20) but overlap
$25$--$86$/100 at top-100 on seven other models (Appendix~\ref{app:visual-verification}).

\begin{table}[h]
\centering
\begin{minipage}[t]{0.485\linewidth}
\centering\vspace{0pt}
\resizebox{\linewidth}{!}{%
\begin{tabular}{lccc}
\toprule
What changes & Pairs & Shared (of 100) & Of ceiling \\
\midrule
nothing (two random halves) & 10 & $96$--$99$ & $1.00$ \\
corpus & 5 & $84$ [$78$--$91$] & $0.86$ \\
query type & 4 & $55$ [$35$--$79$] & $0.56$ \\
attention source & 8 & $49$ [$36$--$79$] & $0.50$ \\
\bottomrule
\end{tabular}}
\captionof{table}{How much the top-100 set changes with each discovery choice: median [range]
over pairs of settings at 200 strips per half; ``of ceiling'' divides by the split-half overlap
(Appendix~\ref{app:noise-ceiling}).}
\label{tab:noise-main}
\end{minipage}\hfill
\begin{minipage}[t]{0.485\linewidth}
\centering\vspace{0pt}
\resizebox{\linewidth}{!}{%
\begin{tabular}{lccc}
\toprule
& \multicolumn{3}{c}{overlap with native set} \\
\cmidrule(lr){2-4}
Query + attention source & ord-last & ref-qspan & ref-out \\
\midrule
ordinal + last (\vir) & $\mathbf{19}$ & $2$ & $5$ \\
ordinal + query span & $7$ & $10$ & $14$ \\
referring + query span (\virrq) & $2$ & $\mathbf{13}$ & $5$ \\
referring + last & $5$ & $5$ & $7$ \\
\bottomrule
\end{tabular}}
\captionof{table}{Top-20 overlap (of 20) between crossover head-sets on the same 200 COCO strips
and each native head-set. Bold: the row's own native score (Appendix~\ref{app:d17}).}
\label{tab:crossover-main}
\end{minipage}
\end{table}

\subsection{The \texorpdfstring{$2\times3\times2$}{2x3x2} design: necessity and steerability}
\label{sec:ncs-main-result}

\begin{figure}[t]
\centering
\includegraphics[width=\linewidth]{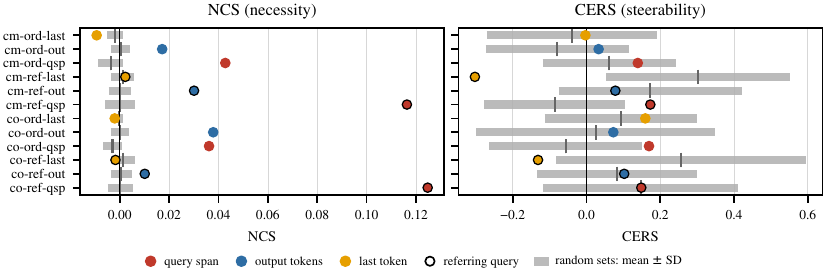}
\vspace{-5mm}
\caption{Table~\ref{tab:ncs-vs-cers} at a glance: each discovered head-set (dot) against the mean
$\pm$ SD of its 20 layer-matched random sets (bar). On NCS the query-span and output-token sets
lie outside their random bands, which sit at zero; on \cers\ every set falls inside or below a wide
random band.}
\label{fig:ncs-vs-cers}
\vspace{-2mm}
\end{figure}

All 12 head-sets (flat top-100) are scored at $N{=}400$ against 20 layer-matched random
head-sets each (Table~\ref{tab:ncs-vs-cers}, Figure~\ref{fig:ncs-vs-cers}). \cers\ uses a $2\times2$ grid of ImageNet photos
and NCS the ImageNet MCQ; ImageNet is never used for discovery.

\begin{table}[t]
\centering
\scriptsize
\setlength{\tabcolsep}{2.2pt}
\renewcommand{\arraystretch}{0.95}
\resizebox{\linewidth}{!}{%
\begin{tabular}{lccccccc}
\toprule
& \multicolumn{4}{c}{\cers\ (steerability)} & \multicolumn{3}{c}{NCS (necessity)} \\
\cmidrule(lr){2-5}\cmidrule(lr){6-8}
& $S$ & layer-matched random & & any-layer random & $S$ & layer-matched random & \\
Setting & [95\% CI] & mean [95\% CI] & $p$ & mean [95\% CI] & [95\% CI] & mean [95\% CI] & $p$ \\
\midrule
cm-ord-last & $-$.003 [$-$.033, .027] & $-$.039 [$-$.147, .069] & .43 & .268 [.171, .364] & $-$.009 [$-$.021, .003] & $-$.002 [$-$.003, .000] & 1.0 \\
cm-ord-out & .033 [.002, .064] & $-$.079 [$-$.170, .011] & .24 & $-$.028 [$-$.140, .084] & .017 [$-$.010, .044] & .000 [$-$.002, .002] & .048 \\
cm-ord-qsp & .139 [.111, .167] & .061 [$-$.023, .146] & .33 & .158 [.018, .297] & .043 [.011, .075] & $-$.004 [$-$.006, $-$.001] & .048 \\
cm-ref-last & $-$.303 [$-$.341, $-$.264] & .303 [.186, .420] & 1.0 & .136 [.005, .267] & .002 [$-$.029, .033] & .001 [$-$.001, .003] & .38 \\
cm-ref-out & .078 [.053, .103] & .173 [.057, .290] & .62 & .149 [.019, .279] & .030 [$-$.001, .061] & .000 [$-$.002, .002] & .048 \\
cm-ref-qsp & .173 [.130, .216] & $-$.086 [$-$.176, .003] & .14 & .060 [$-$.082, .203] & \textbf{.116} [.078, .155] & .002 [.001, .003] & .004$^\ast$ \\
co-ord-last & .159 [.128, .191] & .092 [$-$.004, .189] & .38 & .182 [.089, .275] & $-$.002 [$-$.008, .004] & .000 [$-$.001, .001] & .86 \\
co-ord-out & .073 [.037, .108] & .025 [$-$.127, .176] & .43 & $-$.006 [$-$.129, .117] & .038 [.005, .071] & .000 [$-$.002, .002] & .048 \\
co-ord-qsp & .169 [.138, .201] & $-$.056 [$-$.154, .041] & .24 & .104 [.002, .207] & .036 [.002, .070] & $-$.003 [$-$.005, $-$.001] & .048 \\
co-ref-last & $-$.132 [$-$.154, $-$.109] & .257 [.098, .415] & .81 & .189 [.042, .336] & $-$.002 [$-$.035, .032] & .001 [$-$.001, .004] & .76 \\
co-ref-out & .102 [.079, .125] & .084 [$-$.018, .185] & .43 & .331 [.203, .459] & .010 [$-$.018, .039] & .001 [$-$.001, .003] & .095 \\
co-ref-qsp & .148 [.114, .183] & .147 [.024, .271] & .43 & .132 [$-$.006, .270] & \textbf{.126} [.086, .165] & $-$.001 [$-$.002, $-$.001] & .004$^\ast$ \\
\bottomrule
\end{tabular}}
\caption{All 12 settings, Qwen3-VL-8B, flat top-100 heads, $N{=}400$ (cm/co = comics/COCO corpus;
ord/ref = query; last/qsp/out = attention source). $S$: the discovered head-set, with a 95\%
$t$-interval over the 400 items. Random: mean over 20 random head-sets, either layer-matched or
drawn from any layer, with a 95\% $t$-interval of that mean over the draws.
$p$ = rank $p$ of $S$ among the individual layer-matched sets, whose spread is wider than the
interval of their mean. $^\ast$250 layer-matched random sets (Holm-adjusted $p{=}0.048$).}
\label{tab:ncs-vs-cers}
\end{table}

\begin{table}[t]
\begin{minipage}[t]{0.40\linewidth}
\centering\vspace{0pt}
\scriptsize
\setlength{\tabcolsep}{3pt}
\resizebox{\linewidth}{!}{%
\begin{tabular}{lcc}
\toprule
NCS check, referring + query span & cm & co \\
\midrule
discovered set (top-100) & $.116$ & $.126$ \\
\quad 250 random sets: mean & $.002$ & $-.001$ \\
\quad effect in control SDs & $17.5$ & $19.2$ \\
top-20 heads (above all 20 random) & $.036$ & $.015$ \\
top-160 heads (above all 20 random) & $.160$ & $.145$ \\
generic-damage sets (App.~\ref{app:e2-stress}) & -- & $\le.025$ \\
top-100 image-attention heads & -- & $-.002$ \\
Qwen3-VL-4B / Qwen2-VL-7B & -- & $.054$ / $.029$ \\
\bottomrule
\end{tabular}}
\captionof{table}{Necessity checks on the referring + query-span sets (cm/co = comics/COCO
discovery): the discovered set against 250 layer-matched random sets (rows 1--3), at 20 and 160
heads (4--5), generic-damage and image-heavy head-sets (6--7), and two further models (8).
The discovered sets beat every random set in rows 1--5 and 8; the damage-control sets in rows 6--7
stay near zero (Appendix~\ref{app:twelvecell-e1}, \ref{app:e3-ksweep}, \ref{app:e2-stress}, \ref{app:second-model}).}
\label{tab:necessity-checks}
\end{minipage}\hfill
\begin{minipage}[t]{0.57\linewidth}
\centering\vspace{0pt}
\includegraphics{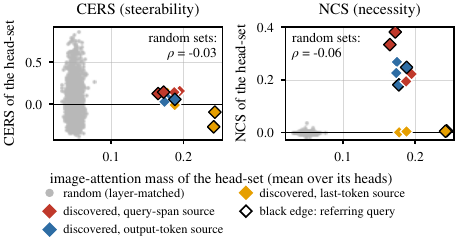}
\captionof{figure}{Image-attention mass does not explain steerability or necessity. $x$: mean
fraction of the heads' attention on image tokens; $y$: \cers\ (left) or NCS (right). Grey: 1{,}200
layer-matched random sets (Appendix~\ref{app:mass-cers}); diamonds: discovered sets. Random \cers\
varies widely at the same mass, and random NCS overlaps at ${\approx}0$. Discovered sets carry
3--6$\times$ the mass yet steer within the random range; last-token sets are as image-heavy as
query-span sets but not necessary.}
\label{fig:mass-scores}
\end{minipage}
\end{table}

\begin{table}[t]
\centering
\scriptsize
\setlength{\tabcolsep}{3.5pt}
\begin{tabular}{lccccc}
\toprule
& \multicolumn{3}{c}{flat top-100: NCS when masked at} & \multicolumn{2}{c}{top-3 heads per layer (108)} \\
\cmidrule(lr){2-4}\cmidrule(lr){5-6}
Setting & all rows & question/option rows & last row & NCS & \cers\ / random ($p$) \\
\midrule
co-ref-qsp & $\mathbf{.126}$ [.086, .165] & $\mathbf{.140}$ [.100, .181] & .000 [$-$.003, .003] & $\mathbf{.046}$ [.013, .078] & .229 / .103 (.38) \\
co-ord-qsp & $\mathbf{.036}$ [.002, .070] & $\mathbf{.049}$ [.015, .083] & $-$.001 [$-$.005, .002] & $\mathbf{.031}$ [$-$.004, .066] & .207 / $-$.091 (.29) \\
co-ref-last & $-$.002 [$-$.035, .032] & $-$.001 [$-$.031, .028] & $-$.003 [$-$.007, .001] & $\mathbf{.012}$ [$-$.020, .045] & .190 / .071 (.33) \\
co-ord-last & $-$.002 [$-$.008, .004] & .002 [$-$.003, .006] & $-$.001 [$-$.004, .002] & .005 [$-$.028, .038] & .458 / .195 (.24) \\
\bottomrule
\end{tabular}
\caption{Where and with which heads necessity acts (COCO, $N{=}400$; 95\% CIs; bold: above all
20 layer-matched random sets, whose means lie within $0.003$ of zero). Left: the flat top-100 set
masked from the image at every row, only at the question and option rows, or only at the last
row. Right: the top-3 heads of every layer instead of the flat top-100 (Appendix~\ref{app:position}).}
\label{tab:selection-position}
\end{table}

\begin{table}[t]
\centering
\scriptsize
\setlength{\tabcolsep}{4pt}
\begin{tabular}{lcccc}
\toprule
& \multicolumn{2}{c}{comics discovery} & \multicolumn{2}{c}{COCO discovery} \\
\cmidrule(lr){2-3}\cmidrule(lr){4-5}
Query + source & NCS [95\% CI] & $p$ & NCS [95\% CI] & $p$ \\
\midrule
ord-last & .004 [$-$.001, .009] & .010 & .001 [.000, .001] & .010 \\
ord-out & .226 [.157, .294] & .010 & .268 [.195, .340] & .010 \\
ord-qsp & .194 [.129, .259] & .010 & .222 [.154, .290] & .010 \\
ref-last & .005 [$-$.004, .014] & .11 & .007 [$-$.003, .016] & .050 \\
ref-out & .247 [.176, .317] & .010 & .181 [.117, .244] & .010 \\
ref-qsp & .381 [.301, .461] & .010 & .333 [.257, .409] & .010 \\
\bottomrule
\end{tabular}
\caption{Necessity as a region contrast on the leak-audited grid ($N{=}120$): $P(c)$ with another
cell hidden from $S$ minus $P(c)$ with the target cell hidden, so generic damage cancels within
each item (maximum $0.716$). $p$: rank $p$ against 100 layer-matched random sets, whose means lie
within $0.002$ of zero (floor $0.010$; Appendix~\ref{app:clean-grid}).}
\label{tab:clean-grid-main}
\end{table}

\textbf{Necessity.} All four query-span head-sets have NCS intervals above zero on both
corpora and beat all 20 random sets; no last-token set does, and Gaze Heads' own setting
(ord-last) is null on both corpora. As a factor, with the 400 items as the unit, query-span
sets exceed last-token sets by $0.083$ ($p{=}6{\times}10^{-10}$; ordinal $0.045$, $p{=}0.003$;
referring $0.121$, $p{=}4{\times}10^{-10}$). Per setting, only the two referring + query-span
sets survive Holm correction: they beat all 250 random sets by 17--19 control SDs
(Table~\ref{tab:necessity-checks}), head-sets built to cause generic damage score at most
$0.025$, and the 100 heads with the most image attention score $-0.002$ ($p{=}0.71$ against
layer-matched random sets; Table~\ref{tab:necessity-checks}, Appendix~\ref{app:e2-stress}), so the effect is neither generic damage
from masking 9\% of the heads nor a property of image-heavy heads. The discovered query-span
sets share only 38--48 of their 100 heads with this image-heavy set. The effect is present from 20 heads and larger for bigger sets ($0.015$--$0.045$ at 20--60
heads, $0.09$--$0.20$ at 100--160; Figure~\ref{fig:ksweep-main}), and it is carried by heads that two
settings share:
shared heads are necessary only when discovered from the query span (Appendix~\ref{app:shared-unique};
$+0.035$ averaged per item, $p{=}0.01$),
while the smaller setting-specific remainders (9--65 heads) do not reach significance.
The referring + query-span effect replicates on Qwen3-VL-4B ($0.054$) and Qwen2-VL-7B
($0.029$), both above all 20 random sets; on Qwen2-VL-7B a referring + last-token set is also
weakly necessary ($0.024$), so ``no last-token
set'' holds on the Qwen3-VL models only (Appendix~\ref{app:second-model}). The effect comes from the
question: masking the referring + query-span set only at the question and option rows gives
$0.140$, as much as masking it everywhere, while masking it only at the last row gives $0.000$
(Table~\ref{tab:selection-position}). Selecting the top-3 heads of every layer instead of the flat
top-100 keeps the query-span sets above all random sets at a smaller size ($0.031$--$0.046$).
On a separately built, leak-audited grid where necessity is measured as a contrast between
hiding the target cell and hiding another cell, so generic damage cancels within each item, all
eight query-span and output-token sets remove 25--53\% of the attainable effect ($0.18$--$0.38$)
and all four last-token sets at most $1\%$ ($\le0.007$), each against 100 random sets
(Table~\ref{tab:clean-grid-main}, Appendix~\ref{app:clean-grid}).

\textbf{Steerability.} At the default strength ($\beta{=}10^4$, as in prior work), steering is not
specific to the discovered heads. Random head-sets drawn
from the same layers, or from any layer, span the discovered sets' scores
(Table~\ref{tab:ncs-vs-cers}), and no discovered head-set steers better than its layer-matched
random sets (rank $p$
$0.14$--$1.0$). Single random draws range from $-0.34$ to $+0.84$ (layer-matched) and $-0.36$ to
$+0.77$ (any layer). The published validations each compare against one draw: Gaze Heads against
one set of 100 heads sampled from layers 20--35 \citep{gazeheads2026}, and visual retrieval heads
against one set of 20 heads outside the detected set \citep{vrh2026}. A single draw is one sample
from this distribution and cannot show where a discovered set falls within it. The most necessary
set (co-ref-qsp) steers as its random sets do ($0.148$ vs.\ $0.147$); both referring + last-token
sets push the model \emph{away} from the steered cell ($-0.30$, $-0.13$) while their random sets
do not; and Gaze Heads' own setting is at zero on comics ($-0.003$). Against random sets also
matched on image-attention mass, four ordinal sets beat all 20, but those controls vary as widely
(means $-0.29$ to $+0.43$). The 20 layer-matched random sets of a setting occupy identical layers, yet
their scores span up to $1.1$; the image attention of their heads (mean, median, minimum, maximum
or spread of per-head image-attention mass) does not predict the score (pooled over settings,
$|\rho|\le0.09$, all $p>0.15$; jointly $R^2{=}0.03$, 240 sets; Figure~\ref{fig:mass-scores},
Appendix~\ref{app:mass-cers}),
and discovered sets, whose heads carry 3--6$\times$ the image attention of any random set, steer
within the same range. Steering outcomes at this strength are therefore not explained by the
layers or the image attention of the heads. Necessity separates the discovery settings; steering
at this strength does not.

\vspace{-5pt}
\subsection{Random baselines and the pointing suite}
\label{sec:sanity-check}
\label{sec:main-result}
\label{sec:general-causal-revision}

On a 14-task pointing suite (seven converted benchmarks, and seven grids of four independent,
already-solved real examples scored on the target cell's own answer; $N{=}80$ per task; Appendix~\ref{app:pointing}), three converted tasks were flagged for leaks and are reported separately. On the remaining 11 tasks, \vir\ steers more than \virrq\ ($0.139$ vs.\
$0.081$), but a layer-matched random set reaches $73\%$ of \vir's \cers\ ($0.102$; paired
difference $0.037$, $[0.025,0.049]$). Pooled over all 14 tasks the paired difference is $0.027$. Out of domain, on eight benchmarks sharing no images or
prompts with discovery, the four query-span sets exceed their random draws on both scores
(Table~\ref{tab:ood-main}). On five further models, including three non-Qwen architectures
(Table~\ref{tab:crossmodel-main}, Appendix~\ref{app:crossmodel}), each of the 25 query-span and output-token sets exceeds its
random draw on NCS ($0.07$--$0.40$ against $0.00$--$0.10$), and last-token sets stay at or below
$0.024$ except on Qwen3.5-9B ($0.223$); steering is model-dependent, near zero on Gemma-4, Ovis1.5
and Qwen3.5-9B. On the 14-task pointing suite, \vir-style sets steer above zero on seven of eight
models (Appendix~\ref{app:d11}).
\label{sec:ood}

\begin{table}[tb]
\centering
\begin{minipage}[t]{0.47\linewidth}
\centering\vspace{0pt}
\resizebox{\linewidth}{!}{%
\begin{tabular}{lcccc}
\toprule
& \multicolumn{2}{c}{\cers} & \multicolumn{2}{c}{NCS} \\
\cmidrule(lr){2-3}\cmidrule(lr){4-5}
Head-set & $S$ & random & $S$ & random \\
\midrule
cm-ref-qsp & $\mathbf{0.288}$ & $-0.117$ & $\mathbf{0.225}$ & $0.020$ \\
co-ref-qsp & $0.092$ & $0.009$ & $0.211$ & $0.023$ \\
co-ord-qsp & $0.271$ & $-0.141$ & $0.161$ & $0.011$ \\
cm-ord-qsp & $0.216$ & $-0.100$ & $0.143$ & $0.017$ \\
\bottomrule
\end{tabular}}
\captionof{table}{Out of domain, Qwen3-VL-8B: eight benchmarks sharing no images or prompts
with discovery ($N{=}1280$ \cers, $563$ NCS pairs); random: one layer-matched draw per head-set
(Appendix~\ref{app:d20}).}
\label{tab:ood-main}
\end{minipage}
\hfill
\begin{minipage}[t]{0.49\linewidth}
\centering\vspace{0pt}
\includegraphics{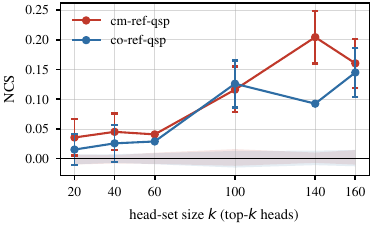}
\captionof{figure}{NCS against head-set size (top-$k$); above all 20 layer-matched random sets
at every $k$ (shaded: random mean $\pm$2 SD; bars: 95\% CI; Appendix~\ref{app:e3-ksweep}).}
\label{fig:ksweep-main}
\end{minipage}
\end{table}

\begin{table}[H]
\centering
\scriptsize
\setlength{\tabcolsep}{4pt}
\caption{Five further models, eight out-of-domain benchmarks (heads discovered anew per model,
flat top-100; one layer-matched random draw per head-set). Ranges over the five query-span and
output-token sets and over their random draws; last token: \texttt{cm-ord-last}
(Appendix~\ref{app:crossmodel}).}
\label{tab:crossmodel-main}
\begin{tabular}{lcccccc}
\toprule
& \multicolumn{4}{c}{NCS (necessity)} & \multicolumn{2}{c}{\cers\ (steerability)} \\
\cmidrule(lr){2-5}\cmidrule(lr){6-7}
& \multicolumn{2}{c}{query span + output} & \multicolumn{2}{c}{last token} & \multicolumn{2}{c}{query span + output} \\
\cmidrule(lr){2-3}\cmidrule(lr){4-5}\cmidrule(lr){6-7}
Model & $S$ & random & $S$ & random & $S$ & random \\
\midrule
Qwen3-VL-4B & $.140$--$.211$ & $.006$--$.032$ & $.010$ & $.002$ & $-.015$--$.154$ & $-.041$--$-.020$ \\
Qwen3.5-9B & $.217$--$.231$ & $.025$--$.043$ & $.223$ & $.060$ & $-.034$--$-.003$ & $-.041$--$.040$ \\
Gemma-4-E4B & $.366$--$.402$ & $.070$--$.097$ & $.022$ & $.035$ & $-.005$--$.007$ & $-.026$--$.038$ \\
Ovis1.5-Llama3-8B & $.204$--$.254$ & $.005$--$.023$ & $.024$ & $.004$ & $.001$--$.014$ & $-.035$--$-.001$ \\
InternVL3.5-14B (4-bit) & $.071$--$.133$ & $.004$--$.033$ & $.024$ & $.012$ & $.228$--$.379$ & $-.176$--$.135$ \\
\bottomrule
\end{tabular}
\end{table}

\vspace{-5mm}
\section{Discussion and Recommendations}
\label{sec:discussion}
\vspace{-5mm}

The firmest result is about the discovery scores themselves: where attention is read from,
and how the question is phrased, shape which heads a score returns, and head-sets discovered
from different positions differ in how much the model needs them. A claim that ``the''
visual routing heads of a model have been found should therefore state the attention source,
query and corpus that produced them. We recommend: \textbf{(i)} compare every head-set against many random draws matched on layer
\emph{and} image-attention mass, and report where it falls among them, since random baselines
for steering are wide; \textbf{(ii)} for necessity, check the score on head-sets built to cause
generic damage; \textbf{(iii)} audit converted benchmarks with oracle, floor, blank-image and
duplicate-option checks; \textbf{(iv)} sweep head-set size.

\vspace{-5mm}\paragraph{Limitations.}
\label{sec:limitations}
The 12-setting design is run on Qwen3-VL-8B, with second-model checks on Qwen3-VL-4B and
Qwen2-VL-7B, out-of-domain results on five further models and pointing-suite results on eight.
Steering is
evaluated at the default strength used in prior work; a pilot at weaker strength
(Appendix~\ref{app:beta}) suggests that several discovered sets separate from random ones there. Masking is attention knockout; mean or
resample ablation and activation patching are natural extensions. The out-of-domain and
cross-model comparisons use one random draw per head-set. Image-attention mass is measured on a generic image prompt, and random head-sets
span a narrower range of it than discovered sets.

\vspace{-2mm}\paragraph{Conclusion.}
The attention source and the query type, far more than
the corpus, shape which heads are found. Head-sets discovered from query-span attention are
more necessary than those discovered from last-token attention, and the two referring +
query-span head-sets are the only ones individually above random controls after
correction. Steering at the default strength, by contrast, is matched by random head-sets drawn
from the same or from any layers, so the two standard validations answer different questions:
necessity separates the discovery settings, while steering at that strength does not. We hope the recommendations above make future
claims about visual heads easier to check.

\vspace{-5pt}
\subsection*{AI use statement}
An AI coding assistant (Claude, Anthropic) assisted with writing and running parts of the experiment code, producing figures, and correcting grammar. No generative AI was used to create synthetic data, develop the core theory, mathematical claims, or proofs.
\vspace{-5pt}
\subsection*{Ethics statement}
This work analyzes internal attention mechanisms of publicly released VLMs using public
benchmark data; no human-subjects data was collected. Identifying and ablating visual heads
could in principle be used to suppress a model's grounding in specific image regions; we see
no realistic misuse path beyond what open-weight attention access already permits.
\vspace{-5pt}
\subsection*{Reproducibility statement}
Code, discovery scores, per-example outputs and the corrected benchmark conversions will be
released with the camera-ready version.
Appendix~\ref{app:protocol} lists the settings behind every table, and
Appendix~\ref{app:algorithm} gives the selection algorithm.

\bibliography{references_related}
\bibliographystyle{iclr2027_conference}
\newpage
\appendix

\appendix

\section*{Overview of the appendix}

\begin{itemize}
\item \textbf{Appendix~\ref{app:twelvecell-full}, the 12-setting design:} the settings; the two
random-control types and the $R{=}250$ result; the two arms of NCS; generic-damage and
image-heavy head-sets; masking restricted to groups of rows and per-layer selection; head-set
size; shared and setting-specific heads; the leak-audited grid; image attention and steering;
and the second-model check.
\item \textbf{Appendix~\ref{app:discovery}, discovery overlap:} the noise ceiling, the crossover
of query and attention source, and top-100 overlap on other models.
\item \textbf{Appendix~\ref{app:d20}, out of domain:} the baseline audit, \cers\ and NCS on eight
benchmarks, and the same evaluation on five further models (full values behind
Table~\ref{tab:crossmodel-main}).
\item \textbf{Appendix~\ref{app:pointing}, the 14-task pointing suite:} its construction, the
discovered set against a random set, and steering on eight further models.
\item \textbf{Appendix~\ref{app:method-details}, method details:} an illustration of \cers\ and
NCS, the per-layer selection algorithm, the protocol behind every result, and the benchmark
conversion recipes.
\end{itemize}

\section{The 12-setting design: settings and controls}
\label{app:twelvecell-full}

\subsection{Settings}
\label{app:twelvecell-method}

A discovery score (Eq.~\ref{eq:discovery}) bundles three choices, which we cross fully
($2\times3\times2=12$ settings) instead of comparing only the two published native settings.
Every level is borrowed from Gaze Heads \citep{gazeheads2026} or visual retrieval heads
\citep{vrh2026}, not chosen by us:

\begin{itemize}
\item \textbf{Query type} (2 levels): \emph{ordinal} (\textsc{ord}, ``What is happening in the
$N$-th panel from the left?'') or \emph{referring} (\textsc{ref}, ``Find the truck.''). Ordinal
queries address a position in a known layout; referring queries address an object by
description, with no positional cue. Ordinal is the Gaze Heads query, referring the
visual-retrieval-head query.
\item \textbf{Attention source} $Q_x$ (3 levels): the last prompt token (\textsc{last}), the
query-span tokens (\textsc{qsp}), or the generated-answer tokens (\textsc{out}), i.e.\ which
token positions' attention to the target region is averaged in $M_{\ell,h}$. Gaze Heads read
the last token; visual retrieval heads read the query span or the output tokens.
\item \textbf{Discovery corpus} (2 levels): $1\times6$ horizontal strips of comic panels
(\textsc{cm}) or COCO photos (\textsc{co}), each with one uniquely identified target panel, so
both query types are well posed on both corpora. Comics is the Gaze Heads discovery corpus;
COCO photos with object boxes are the grounding data of our visual-retrieval-head
reproduction.
\end{itemize}

The steering strength $\beta{=}10^4$ and the boost/suppress form of Eq.~\ref{eq:steer} are the
defaults of the released Gaze Heads code. Settings are named \texttt{<corpus>-<query>-<source>}, e.g.\ \texttt{cm-ref-qsp} = comics
corpus, referring query, query-span source. Gaze Heads' setting is \texttt{*-ord-last} (\vir);
the visual-retrieval-head score is \texttt{*-ref-qsp} (\virrq) or \texttt{*-ref-out}
(\virro). All results in this section use Qwen3-VL-8B-Instruct, flat top-100 heads and the same
$N{=}400$ evaluation items unless stated.

\subsection{Random controls}
\label{app:twelvecell-e1}

For each setting we draw $R{=}20$ random head-sets of two kinds, both with the discovered set's
exact per-layer head counts:

\begin{itemize}
\item \textbf{Control A (layer-matched):} each head is drawn uniformly from the same layer as
the discovered head it replaces, excluding the discovered set's own heads.
\item \textbf{Control B (layer- and attention-matched):} as Control A, but each head is also
drawn from the same quintile of image-attention mass as the head it replaces (mean attention
on image tokens at the last prompt token over $200$ unrelated single-image forward passes).
Draws are without replacement and exclude the discovered heads; when a (layer, quintile) bin
is exhausted, the nearest quintile in the same layer is used.
\item \textbf{Any-layer control} (\cers\ only, the ``any-layer random'' column of
Table~\ref{tab:ncs-vs-cers}): $100$ heads drawn without replacement from all heads outside the
discovered set, ignoring layer.
\end{itemize}

The $p$-value is the rank $p=(b+1)/(R+1)$, where $b$ is the number of controls scoring at
least as high as the discovered set (floor $1/21\approx0.048$ at $R{=}20$). Scores and
layer-matched means with 95\% intervals are in Table~\ref{tab:ncs-vs-cers};
Table~\ref{tab:e1-random-control} adds the attention-matched controls.

\begin{table}[h]
\centering
\scriptsize
\caption{Rank $p$ of each discovered head-set against 20 layer-matched (A) and 20 layer- and
attention-matched (B) random head-sets (Qwen3-VL-8B, flat top-100, $N{=}400$). Scores are in
Table~\ref{tab:ncs-vs-cers}.}
\label{tab:e1-random-control}
\begin{tabular}{lcccc}
\toprule
& \multicolumn{2}{c}{\cers} & \multicolumn{2}{c}{NCS} \\
\cmidrule(lr){2-3}\cmidrule(lr){4-5}
Setting & A & B & A & B \\
\midrule
cm-ord-last & .43 & .095 & 1.00 & 1.00 \\
cm-ord-out & .24 & .048 & .048 & .048 \\
cm-ord-qsp & .33 & .048 & .048 & .048 \\
cm-ref-last & 1.00 & 1.00 & .38 & .24 \\
cm-ref-out & .62 & .24 & .048 & .048 \\
cm-ref-qsp & .14 & .095 & .048 & .048 \\
co-ord-last & .38 & .86 & .86 & .76 \\
co-ord-out & .43 & .048 & .048 & .048 \\
co-ord-qsp & .24 & .048 & .048 & .048 \\
co-ref-last & .81 & .81 & .76 & .76 \\
co-ref-out & .43 & .76 & .095 & .048 \\
co-ref-qsp & .43 & .86 & .048 & .048 \\
\bottomrule
\end{tabular}
\end{table}

\textbf{Necessity.} All four query-span sets and three of the four output-token sets beat all
20 draws of both control types. The four last-token sets fail both ($p\ge0.24$), including Gaze
Heads' own setting on both corpora (\texttt{cm-ord-last} $p{=}1.00$, \texttt{co-ord-last}
$p{=}0.86$). The two referring + last-token sets share the referring query with the strongest
sets and still fail, so the attention source, not the query wording, carries the effect.
\texttt{co-ref-out} beats all attention-matched draws but not all layer-matched ones.
Attention matching therefore removes none of the necessity effects.

\textbf{Steerability.} No set beats its layer-matched controls (rank $p$ $0.14$--$1.00$);
individual draws range from $-0.34$ to $+0.84$ over the 240 draws. Four ordinal sets beat all
20 attention-matched draws, but those controls vary as widely as the layer-matched ones
(means $-0.29$ to $+0.43$).

\textbf{Multiple comparisons.} At $R{=}20$ the smallest attainable $p$ is $0.048$, which Holm
correction over 12 settings cannot pass. We therefore reran the two referring + query-span
settings with $R{=}250$ layer-matched controls (floor $1/251\approx0.004$;
Table~\ref{tab:e1-r250}). Both beat all 250 controls, $17.5$ and $19.2$ control standard
deviations above the control mean, and both survive Holm correction over the 12 settings
(unadjusted $p{=}1/251\approx0.004$; Holm-adjusted $p{=}0.048$ for both,
since $12\times0.004=0.048$ and the next Holm step is
$\max(0.048,11\times0.004)=0.048$); the other ten settings are bounded by the
$R{=}20$ floor.

\begin{table}[h]
\centering
\scriptsize
\caption{The two referring + query-span settings against $R{=}250$ layer-matched random
head-sets (Qwen3-VL-8B, top-100, $N{=}400$): control mean $\pm$ SD over the 250 draws and the
rank $p$ (floor $0.004$).}
\label{tab:e1-r250}
\begin{tabular}{lccc}
\toprule
Setting & NCS & Control mean $\pm$ SD & $p$ \\
\midrule
cm-ref-qsp & $0.116$ & $0.002\pm.007$ & $.004$ \\
co-ref-qsp & $0.126$ & $-0.001\pm.007$ & $.004$ \\
\bottomrule
\end{tabular}
\end{table}

\subsection{The two arms of NCS}
\label{app:d19}

Table~\ref{tab:ncs-arm-audit} splits each NCS value into its two arms, using an independent
$N{=}400$ replicate of the 12 settings (same head-sets; values differ from
Table~\ref{tab:ncs-vs-cers} by at most $0.006$). The unmasked baselines are shared across
settings: $P(\text{correct})=0.995$ ($[0.988,0.999]$) for matched images and $0.217$
($[0.185,0.252]$) for mismatched images. The mismatched arm is bimodal per example (median
$0.007$; $62\%$ of examples below $0.05$, $20\%$ above $0.5$), and masking moves individual
mismatched examples (mean absolute change $0.14$--$0.21$ in the non-\textsc{last} settings),
but these moves cancel: the mean mismatched drop is within $\pm0.013$ of zero, with an interval
including zero, in every setting. NCS therefore equals the matched drop up to noise, with an
interval $1.5$--$7\times$ wider (median $2.2\times$).

\begin{table}[h]
\centering
\scriptsize
\caption{NCS split into its arms, all 12 settings ($N{=}400$, top-100). $P$ renormalized over
the four letters; drops are baseline minus masked. $95\%$ percentile bootstrap intervals,
$2{,}000$ resamples.}
\label{tab:ncs-arm-audit}
\begin{tabular}{lccccc}
\toprule
Setting & $P_{\text{mask}}$ matched & Matched drop & $P_{\text{mask}}$ mism. & Mism.\ drop & NCS \\
\midrule
cm-ord-last & $0.996$ & $-0.002$ {\tiny[-0.005,0.001]} & $0.212$ & $0.005$ {\tiny[-0.006,0.017]} & $-0.007$ {\tiny[-0.019,0.004]} \\
cm-ord-out & $0.973$ & $0.022$ {\tiny[0.009,0.036]} & $0.213$ & $0.005$ {\tiny[-0.019,0.028]} & $0.017$ {\tiny[-0.010,0.045]} \\
cm-ord-qsp & $0.967$ & $0.028$ {\tiny[0.015,0.042]} & $0.230$ & $-0.013$ {\tiny[-0.042,0.018]} & $0.040$ {\tiny[0.009,0.070]} \\
cm-ref-last & $0.990$ & $0.004$ {\tiny[-0.000,0.010]} & $0.217$ & $0.000$ {\tiny[-0.030,0.031]} & $0.004$ {\tiny[-0.027,0.034]} \\
cm-ref-out & $0.963$ & $0.032$ {\tiny[0.017,0.048]} & $0.216$ & $0.002$ {\tiny[-0.025,0.030]} & $0.030$ {\tiny[-0.001,0.059]} \\
cm-ref-qsp & $0.885$ & $0.110$ {\tiny[0.083,0.135]} & $0.220$ & $-0.003$ {\tiny[-0.034,0.029]} & $0.113$ {\tiny[0.074,0.153]} \\
co-ord-last & $0.995$ & $-0.001$ {\tiny[-0.002,0.000]} & $0.217$ & $0.000$ {\tiny[-0.006,0.005]} & $-0.001$ {\tiny[-0.006,0.005]} \\
co-ord-out & $0.956$ & $0.039$ {\tiny[0.022,0.056]} & $0.218$ & $-0.000$ {\tiny[-0.028,0.028]} & $0.039$ {\tiny[0.007,0.071]} \\
co-ord-qsp & $0.960$ & $0.035$ {\tiny[0.020,0.051]} & $0.219$ & $-0.002$ {\tiny[-0.031,0.029]} & $0.036$ {\tiny[0.002,0.069]} \\
co-ref-last & $0.987$ & $0.008$ {\tiny[0.001,0.017]} & $0.208$ & $0.009$ {\tiny[-0.023,0.042]} & $-0.001$ {\tiny[-0.035,0.031]} \\
co-ref-out & $0.984$ & $0.011$ {\tiny[0.001,0.021]} & $0.216$ & $0.001$ {\tiny[-0.025,0.028]} & $0.010$ {\tiny[-0.019,0.038]} \\
co-ref-qsp & $0.883$ & $0.112$ {\tiny[0.088,0.136]} & $0.230$ & $-0.013$ {\tiny[-0.045,0.020]} & $0.125$ {\tiny[0.086,0.163]} \\
\bottomrule
\end{tabular}
\end{table}

\subsection{Generic-damage and image-heavy head-sets}
\label{app:e2-stress}

Because the mismatched arm sits at chance, it cannot cancel generic damage
(\S\ref{sec:necessity-method}), so a positive NCS could in principle come from any head-set
whose masking disrupts the forward pass. We tested this on \texttt{co-ref-qsp} against five
head-sets built to cause generic damage: the $100$ heads with the most attention on the first
prompt token (attention sinks), the $100$ with the most attention on text tokens, all heads of
three contiguous layers ($96$ heads), and uniformly random sets of $200$ and $400$ heads. NCS-v2
replaces the mismatched arm with a text-solvable multiple-choice question shown with an
irrelevant photo: masking a head-set's image attention should hurt the matched arm and leave
this arm untouched. To check that this arm registers generic damage, we also masked each
head-set's attention to \emph{text} on it.

\begin{table}[h]
\centering
\scriptsize
\caption{Generic-damage controls on \texttt{co-ref-qsp} ($N{=}400$). Matched and mismatched drops
are the decrease in $P(\text{correct})$ when the head-set's image attention is masked; the last
column is the decrease on the text-solvable arm when the head-set's text attention is masked.}
\label{tab:e2-stress}
\begin{tabular}{lcccccc}
\toprule
Head-set & Heads & Matched drop & Mismatched drop & NCS & NCS-v2 & Text-arm drop (text masked) \\
\midrule
Discovered (\virrq) & 100 & $0.112$ & $-0.013$ & $\mathbf{0.125}$ & $\mathbf{0.112}$ & $0.645$ \\
Attention sinks & 100 & $0.002$ & $-0.005$ & $0.007$ & $0.002$ & $0.012$ \\
Text-heavy heads & 100 & $0.002$ & $-0.003$ & $0.005$ & $0.002$ & $0.415$ \\
Whole layers & 96 & $0.005$ & $0.006$ & $-0.001$ & $0.005$ & $0.232$ \\
Random, 200 heads & 200 & $0.006$ & $-0.014$ & $0.020$ & $0.006$ & $0.718$ \\
Random, 400 heads & 400 & $0.010$ & $-0.015$ & $0.025$ & $0.010$ & $0.776$ \\
\bottomrule
\end{tabular}
\end{table}

Every generic-damage head-set scores NCS $-0.001$ to $0.025$ and NCS-v2 $0.002$--$0.010$,
against $0.125$ and $0.112$ for the discovered set, although masking the text attention of all
but the sink heads lowers accuracy on the text-solvable arm by $0.23$--$0.78$: these head-sets
can cause damage, but masking their image attention does not remove the answer. Masking the
discovered set's image attention leaves the text-solvable arm untouched ($<10^{-5}$), so NCS-v2
equals its matched drop.

\textbf{Image-heavy heads.} The 100 heads with the most attention on image tokens (the mass
measure of Control B) score NCS $-0.002$ ($[-0.030, 0.026]$) against a layer-matched control
mean of $-0.001$ ($p{=}0.71$, $R{=}20$) and an attention-matched mean of $0.002$ ($p{=}0.76$).
They share $38$--$48$ of their $100$ heads with the four query-span sets and $61$--$63$ with
the two referring + last-token sets.

\subsection{Where necessity acts: masked rows and per-layer selection}
\label{app:position}

Table~\ref{tab:selection-position} (left) restricts the image mask to one group of attention
rows (query positions) of the discovered set: all rows; the text rows after the image except
the final one (question, options and chat template); or only the final prompt row, which
predicts the answer letter. The same 400 items, NCS definition and 20 layer-matched random sets
are used. The right half replaces the flat top-100 with the top-3 heads of every layer
($k_{\mathrm{per\_layer}}{=}3$, 108 heads; Appendix~\ref{app:algorithm}), scored for NCS and
\cers\ against 20 random sets with the same per-layer counts. Table~\ref{tab:position-full}
gives every row group for both selections, including the image rows (image tokens attending to
the image), which Table~\ref{tab:selection-position} omits.

\begin{table}[h]
\centering
\scriptsize
\setlength{\tabcolsep}{2.5pt}
\caption{NCS when the image mask is applied only at one group of attention rows (COCO settings,
$N{=}400$). $S$ with 95\% $t$-interval; random: mean of 20 random head-sets with the same
per-layer counts; $p$: rank $p$ (floor $0.048$).}
\label{tab:position-full}
\begin{tabular}{llcccccc}
\toprule
& & \multicolumn{3}{c}{flat top-100} & \multicolumn{3}{c}{top-3 per layer (108)} \\
\cmidrule(lr){3-5}\cmidrule(lr){6-8}
Setting & Masked rows & $S$ [95\% CI] & random & $p$ & $S$ [95\% CI] & random & $p$ \\
\midrule
co-ref-qsp & all rows & .126 [.086, .165] & .000 & .048 & .046 [.013, .078] & .001 & .048 \\
 & image rows & $-$.001 [$-$.023, .020] & .002 & .86 & .010 [$-$.012, .033] & .000 & .048 \\
 & question/option rows & .140 [.100, .181] & $-$.004 & .048 & .056 [.025, .088] & $-$.001 & .048 \\
 & last row & .000 [$-$.003, .003] & .000 & .67 & $-$.002 [$-$.006, .002] & .000 & 1.0 \\
\midrule
co-ord-qsp & all rows & .036 [.002, .070] & $-$.002 & .048 & .031 [$-$.004, .066] & .003 & .048 \\
 & image rows & $-$.001 [$-$.016, .013] & .000 & .67 & .001 [$-$.018, .020] & $-$.001 & .29 \\
 & question/option rows & .049 [.015, .083] & $-$.003 & .048 & .047 [.013, .082] & .001 & .048 \\
 & last row & $-$.001 [$-$.005, .002] & .000 & .95 & $-$.003 [$-$.007, .002] & .000 & 1.0 \\
\midrule
co-ref-last & all rows & $-$.002 [$-$.035, .032] & .000 & .67 & .012 [$-$.020, .045] & .000 & .048 \\
 & image rows & $-$.002 [$-$.013, .009] & $-$.001 & .57 & $-$.001 [$-$.017, .015] & $-$.002 & .43 \\
 & question/option rows & $-$.001 [$-$.031, .028] & .000 & .76 & .006 [$-$.023, .034] & .001 & .19 \\
 & last row & $-$.003 [$-$.007, .001] & .000 & 1.0 & $-$.003 [$-$.007, .001] & .000 & 1.0 \\
\midrule
co-ord-last & all rows & $-$.002 [$-$.008, .004] & .000 & .95 & .005 [$-$.028, .038] & $-$.001 & .19 \\
 & image rows & .000 [$-$.002, .002] & .000 & .43 & $-$.002 [$-$.013, .008] & $-$.002 & .62 \\
 & question/option rows & .002 [$-$.003, .006] & .000 & .10 & .013 [$-$.019, .045] & $-$.001 & .048 \\
 & last row & $-$.001 [$-$.004, .002] & .000 & .95 & $-$.003 [$-$.006, .001] & .000 & 1.0 \\
\bottomrule
\end{tabular}
\end{table}

For both query-span sets the question and option rows carry the whole effect ($0.140$ and
$0.049$ flat, at least as large as masking all rows), while masking at the image rows or the
last row alone leaves NCS within $0.003$ of zero. With top-3-per-layer selection the pattern is
the same at a smaller size (question and option rows $0.056$ and $0.047$; image rows $0.010$ and
$0.001$), and the last-token sets stay near zero in every row group
($\le0.013$).

\subsection{Head-set size}
\label{app:e3-ksweep}

We sweep $k\in\{20,40,60,100,140,160\}$ on the two referring + query-span settings with the
protocol of Appendix~\ref{app:twelvecell-e1} ($R{=}20$, $N{=}400$; Table~\ref{tab:ksweep},
Figure~\ref{fig:ksweep-main}). NCS is above all layer-matched controls at every $k$ on both
corpora. It rises from $0.015$--$0.045$ at $20$--$60$ heads to $0.09$--$0.20$ at $100$--$160$,
while the control means stay within $0.002$ of zero.

\begin{table}[h]
\centering
\scriptsize
\caption{NCS against head-set size (Qwen3-VL-8B, $N{=}400$). $S$ with 95\% $t$-interval where the
per-example outputs are available; random: mean $\pm$ SD of 20 layer-matched random sets (250 at
$k{=}100$); $p$: rank $p$.}
\label{tab:ksweep}
\begin{tabular}{rcccccc}
\toprule
& \multicolumn{3}{c}{\texttt{cm-ref-qsp}} & \multicolumn{3}{c}{\texttt{co-ref-qsp}} \\
\cmidrule(lr){2-4}\cmidrule(lr){5-7}
$k$ & $S$ [95\% CI] & random & $p$ & $S$ [95\% CI] & random & $p$ \\
\midrule
20 & .036 [.005, .066] & $-$.002 $\pm$ .004 & .048 & .015 [$-$.011, .041] & $-$.001 $\pm$ .004 & .048 \\
40 & .045 [.014, .076] & $-$.001 $\pm$ .003 & .048 & .026 [$-$.005, .056] & .000 $\pm$ .004 & .048 \\
60 & .041 & .001 $\pm$ .005 & .048 & .029 & .000 $\pm$ .005 & .048 \\
100 & .116 [.078, .155] & .002 $\pm$ .007 & .004 & .126 [.086, .165] & $-$.001 $\pm$ .007 & .004 \\
140 & .204 [.160, .249] & .002 $\pm$ .004 & .048 & .092 & .001 $\pm$ .006 & .048 \\
160 & .160 [.119, .202] & .002 $\pm$ .006 & .048 & .145 [.104, .186] & .001 $\pm$ .007 & .048 \\
\bottomrule
\end{tabular}
\end{table}
 \cers\ at $k{=}20$ is $0.175$ (comics) and $0.193$ (COCO), inside its
layer-matched control distribution ($p{=}0.38$, $0.43$), as at $k{=}100$.

\subsection{Shared and setting-specific heads}
\label{app:shared-unique}

\begin{figure}[h]
\centering
\includegraphics[width=\linewidth]{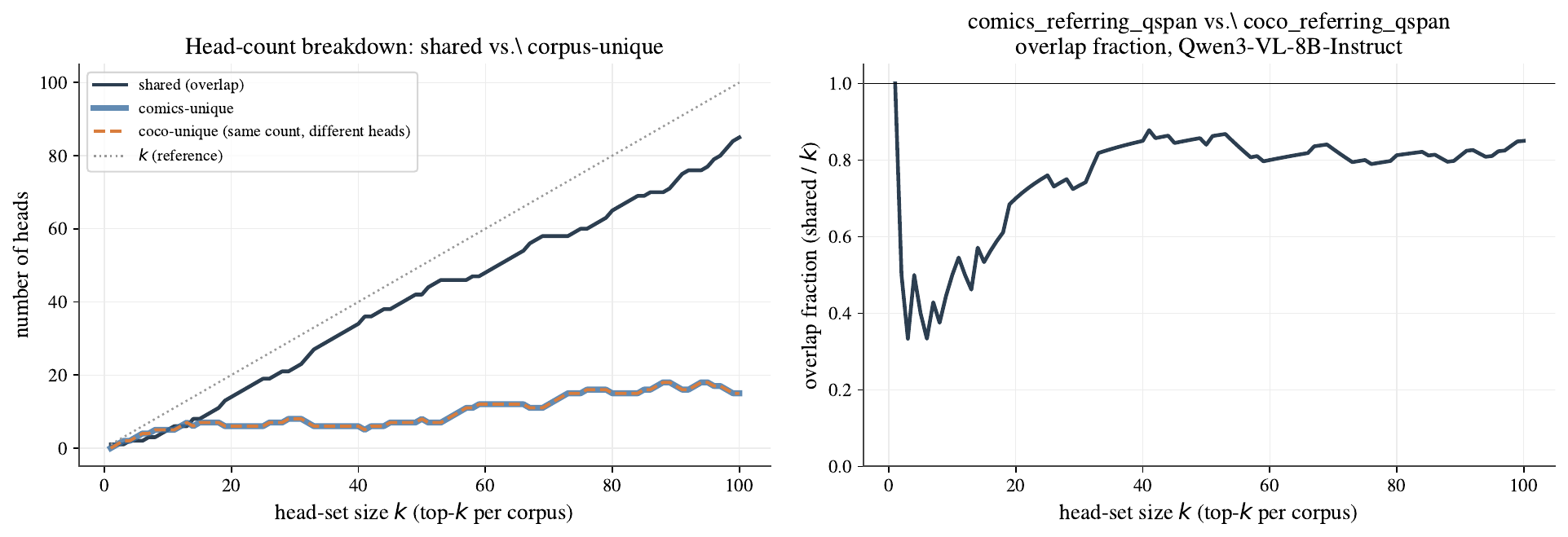}
\caption{Top-$k$ overlap between \texttt{cm-ref-qsp} and \texttt{co-ref-qsp} (Qwen3-VL-8B). Left:
shared heads and each corpus's unique heads (equal in count, not in identity). Right: overlap
fraction (shared$/k$).}
\label{fig:overlap-vs-k}
\end{figure}

\texttt{cm-ref-qsp} and \texttt{co-ref-qsp} share $85$ of their top-100 heads.
Figure~\ref{fig:overlap-vs-k} tracks this overlap as $k$ grows: the two corpora's single top
head is the same, the overlap fraction dips to $33\%$ around $k{\approx}3$--$8$, and it
plateaus near $80$--$85\%$ from $k{\approx}40$. The $15$ corpus-unique heads of each set are
null on NCS against 20 layer-matched controls of the same size: $-0.001$ $[-.008,.006]$
($p{=}.52$) for comics and $0.005$ $[-.010,.021]$ ($p{=}.14$) for COCO.

More generally, we split six pairs of settings into the heads both select (shared) and the heads
only one selects (A-only, B-only), and scored each part's NCS against 20 layer-matched random
controls of the same size. The six pairs form three comparisons, each made once with the query
span and once with the last token as attention source (Table~\ref{tab:shared-unique}). Shared
heads are necessary in all three query-span pairs ($+0.029$ to $+0.036$) and in none of the
three last-token pairs ($-0.003$ to $-0.001$); a paired test over the same items gives a
query-span minus last-token difference of $+0.039$ ($p{=}0.010$), $+0.035$ ($p{=}0.020$) and
$+0.030$ ($p{=}0.035$). The three comparisons share the same 400 items (their per-item
differences correlate at $0.70$--$0.80$), so we average them per item: the mean difference is
$+0.035$ ($[0.008, 0.061]$, $p{=}0.01$, $n{=}400$). None of the $12$ setting-specific sets
shows a necessity effect, but they hold only 9--65 heads against 35--91 for the shared sets. In
the three query-span pairs, 10 random subsets of the shared heads of the same size as each
setting-specific set (9--29 heads) are also null (means $-0.004$ to $0.000$; single subsets
$-0.016$ to $+0.009$), and every setting-specific set lies within the range of its subsets: at
these sizes neither shared nor setting-specific heads show a detectable effect, consistent with
the growth of NCS with set size (Appendix~\ref{app:e3-ksweep}).

\begin{table}[h]
\centering
\scriptsize
\caption{NCS above the mean of 20 same-size layer-matched random controls (paired one-sided $p$
in parentheses; bold $p<0.05$) for shared and setting-specific heads (Qwen3-VL-8B, top-100 per
setting, $N{=}400$). ``Shared / only'' gives the head counts.}
\label{tab:shared-unique}
\begin{tabular}{llcccc}
\toprule
Comparison & Source & Shared / only & Shared & A-only & B-only \\
\midrule
corpus (cm vs co), ordinal & query span & $91$ / $9$ & $\mathbf{+0.036}$ (0.01) & $-0.001$ (0.61) & $+0.001$ (0.35) \\
corpus (cm vs co), ordinal & last token & $80$ / $20$ & $-0.003$ (0.84) & $-0.003$ (0.84) & $+0.000$ (0.49) \\
query type (ord vs ref), comics & query span & $79$ / $21$ & $\mathbf{+0.033}$ (0.015) & $+0.010$ (0.2) & $+0.005$ (0.14) \\
query type (ord vs ref), comics & last token & $35$ / $65$ & $-0.001$ (0.67) & $-0.006$ (0.97) & $+0.005$ (0.37) \\
source (X vs out), COCO ordinal & query span & $71$ / $29$ & $\mathbf{+0.029}$ (0.026) & $+0.011$ (0.19) & $-0.004$ (0.83) \\
source (X vs out), COCO ordinal & last token & $51$ / $49$ & $-0.001$ (0.64) & $+0.000$ (0.42) & $+0.013$ (0.17) \\
\bottomrule
\end{tabular}
\end{table}

\subsection{Region-contrast necessity on a leak-audited grid}
\label{app:clean-grid}

To test necessity with a contrast that cancels generic damage within each item, we built a
separate ImageNet $2\times2$ grid set and scored all 12 head-sets on it against 100
layer-matched random sets each. \textbf{Items.} Classes whose name is shared with another
ImageNet class were removed; the rest were split into disjoint groups of four, so no class or
photo appears twice. Each photo is inset in its own $224$-pixel cell of a $448\times448$ canvas,
so every cell maps to its own $7\times7$ block of image tokens. Of the $248$ grids, the $239$
whose four cells the model names correctly when asked by position and whose target question it
answers correctly were kept; $120$ were drawn at random with the target cell and the correct
letter balanced. \textbf{Questions.} \cers\ uses ``Which of the following is shown in this
image?'' with the four grid classes as options, all of them correct (unsteered $P$(target)
$0.23$, chance $0.25$). NCS uses the same question with one correct option, the target class
plus three classes absent from the grid (unsteered accuracy $0.98$; $0.25$ with the target cell
grayed, $0.26$ on a gray image, $0.99$ with another cell grayed). \textbf{Scores.} \cers\ is the
symmetric version of Eq.~\ref{eq:cers}, averaging the correct and the distractor option. NCS is
$P(c\mid \text{hide }\rho(d)\text{ from }S)-P(c\mid\text{hide }\rho(c)\text{ from }S)$, where
hiding adds $-\beta$ to $S$'s attention logits on one cell's tokens only; both arms apply the
same intervention to an equal-sized cell, so generic damage cancels within each item, and
graying the cell instead of masking gives the attainable maximum, $0.716$.

\begin{table}[h]
\centering
\scriptsize
\setlength{\tabcolsep}{2.5pt}
\caption{Region-contrast NCS and symmetric \cers\ on the leak-audited grid ($N{=}120$ items,
Qwen3-VL-8B, flat top-100). $S$: discovered set, with a 95\% $t$-interval over items; random: mean
[95\% CI] over 100 layer-matched random sets; $p$: rank $p$ (floor $0.010$). The maximum attainable
NCS is $0.716$.}
\label{tab:clean-grid}
\begin{tabular}{lcccccc}
\toprule
& \multicolumn{3}{c}{NCS (region contrast)} & \multicolumn{3}{c}{\cers\ (symmetric)} \\
\cmidrule(lr){2-4}\cmidrule(lr){5-7}
Setting & $S$ [95\% CI] & random & $p$ & $S$ [95\% CI] & random & $p$ \\
\midrule
cm-ord-last & {.004} [$-$.001, .009] & .000 [.000, .000] & .010 & $-$.010 [$-$.052, .031] & $-$.040 [$-$.078, $-$.003] & .40 \\
cm-ord-out & {.226} [.157, .294] & $-$.001 [$-$.001, .000] & .010 & .027 [$-$.014, .068] & $-$.102 [$-$.138, $-$.066] & .21 \\
cm-ord-qsp & {.194} [.129, .259] & .000 [.000, .001] & .010 & .141 [.106, .176] & .067 [.030, .104] & .37 \\
cm-ref-last & .005 [$-$.004, .014] & .001 [.001, .002] & .11 & $-$.272 [$-$.325, $-$.220] & .271 [.218, .325] & .99 \\
cm-ref-out & {.247} [.176, .317] & .001 [.000, .001] & .010 & .057 [.026, .087] & .139 [.090, .187] & .64 \\
cm-ref-qsp & {.381} [.301, .461] & $-$.002 [$-$.002, $-$.001] & .010 & .143 [.083, .202] & $-$.086 [$-$.125, $-$.047] & .16 \\
co-ord-last & {.001} [.000, .001] & .000 [.000, .000] & .010 & .134 [.090, .178] & .127 [.080, .175] & .47 \\
co-ord-out & {.268} [.195, .340] & .000 [$-$.001, .000] & .010 & .048 [.001, .095] & $-$.039 [$-$.083, .005] & .32 \\
co-ord-qsp & {.222} [.154, .290] & .000 [$-$.001, .000] & .010 & .156 [.116, .197] & .004 [$-$.040, .047] & .24 \\
co-ref-last & .007 [$-$.003, .016] & .001 [.001, .002] & .050 & $-$.096 [$-$.128, $-$.065] & .188 [.128, .248] & .74 \\
co-ref-out & {.181} [.117, .244] & .000 [.000, .001] & .010 & .084 [.054, .114] & .129 [.093, .165] & .53 \\
co-ref-qsp & {.333} [.257, .409] & .000 [.000, .000] & .010 & .126 [.076, .176] & .083 [.042, .124] & .33 \\
\bottomrule
\end{tabular}
\end{table}

Every query-span and output-token set removes 25--53\% of the attainable effect and exceeds all
100 random sets, whose values lie within $0.002$ of zero. The last-token sets remove at most
$1\%$; two of them formally exceed all random sets because the random spread is nearly zero. No
set steers better than its random sets (rank $p$ $0.16$--$0.99$).

\subsection{Image attention and steering}
\label{app:mass-cers}

For every head-set in Table~\ref{tab:ncs-vs-cers} we took each head's image-attention mass (the
Control-B measure) and summarized the set by the mean, median, minimum, maximum and standard
deviation of this mass over its $100$ heads. The 20 layer-matched random sets of each setting
were regenerated from their fixed seeds (their recomputed mean \cers\ matches
Table~\ref{tab:ncs-vs-cers} in all 12 settings). Table~\ref{tab:mass-cers} gives, per setting,
the discovered set's mean mass, the random sets' mean mass, and the Spearman correlation between
each mass statistic and \cers\ across the 20 random sets. Signs are inconsistent across
settings; one of the $60$ correlations survives Holm correction (\texttt{co-ord-out}, minimum
mass, $\rho{=}0.71$). Pooled over settings after centering each setting, no statistic predicts
\cers\ ($|\rho|\le0.09$, all $p>0.15$), and the five together explain $R^2{=}0.03$ of the
variance. On the leak-audited grid of Appendix~\ref{app:clean-grid} (1{,}200 random sets over 12
settings, the grey points of Figure~\ref{fig:mass-scores}) the result is the same
($|\rho|\le0.03$, $R^2{=}0.006$). Discovered sets have $3$--$6\times$ the mean image-attention
mass of any of their random sets yet steer within the random range.

\begin{table}[h]
\centering
\scriptsize
\setlength{\tabcolsep}{3pt}
\caption{Image-attention mass of head-sets and its relation to \cers. Mass columns: mean over the
set's heads (random: mean [min, max] over 20 layer-matched sets). $\rho$: Spearman correlation
between the mass statistic and \cers\ across the 20 random sets.}
\label{tab:mass-cers}
\begin{tabular}{lcccccccc}
\toprule
& \multicolumn{2}{c}{mean image-attention mass} & \multicolumn{5}{c}{$\rho$(statistic, \cers) over random sets} \\
\cmidrule(lr){2-3}\cmidrule(lr){4-8}
Setting & discovered & random & mean & median & min & max & SD \\
\midrule
cm-ord-last & 0.188 & 0.041 [0.032, 0.046] & $-$.39 & $-$.54 & $-$.22 & $-$.20 & $-$.14 \\
cm-ord-out & 0.174 & 0.052 [0.045, 0.062] & $-$.04 & .12 & $-$.01 & .28 & .17 \\
cm-ord-qsp & 0.187 & 0.046 [0.040, 0.053] & $-$.15 & .02 & $-$.19 & $-$.10 & $-$.29 \\
cm-ref-last & 0.242 & 0.047 [0.040, 0.061] & .39 & .11 & .26 & .37 & .40 \\
cm-ref-out & 0.188 & 0.051 [0.043, 0.061] & $-$.37 & $-$.40 & $-$.21 & .01 & $-$.12 \\
cm-ref-qsp & 0.172 & 0.049 [0.041, 0.056] & $-$.10 & $-$.08 & .13 & $-$.35 & $-$.33 \\
co-ord-last & 0.178 & 0.041 [0.031, 0.048] & .07 & $-$.18 & .12 & .29 & .15 \\
co-ord-out & 0.175 & 0.050 [0.044, 0.056] & .16 & .31 & .71 & $-$.03 & $-$.09 \\
co-ord-qsp & 0.195 & 0.046 [0.042, 0.052] & $-$.19 & .00 & .10 & $-$.14 & $-$.17 \\
co-ref-last & 0.243 & 0.048 [0.041, 0.056] & $-$.26 & $-$.41 & $-$.07 & .14 & .03 \\
co-ref-out & 0.177 & 0.053 [0.046, 0.058] & $-$.50 & $-$.30 & .09 & $-$.21 & $-$.45 \\
co-ref-qsp & 0.165 & 0.056 [0.046, 0.074] & .21 & .02 & .00 & $-$.04 & .07 \\
\midrule
pooled (240 sets, centered) & & & $-$.09 & $-$.09 & .06 & $-$.04 & $-$.07 \\
\bottomrule
\end{tabular}
\end{table}

\subsection{Steering strength (pilot)}
\label{app:beta}

All \cers\ values in the main text use $\beta{=}10^4$. A pilot sweep over
$\beta\in\{1,3,10,30,100,10^3,10^4\}$ on four settings ($N{=}200$ items, 5 layer-matched random
sets each; Table~\ref{tab:beta}) suggests that the overlap between discovered and random sets
is specific to strong steering. At $\beta{=}3$ the two referring + query-span sets steer strongly ($0.869$,
$0.789$) and above all 5 random sets (means $-0.103$ and $0.204$), and \texttt{cm-ord-last} is
also above all 5 ($0.308$ vs.\ $-0.134$); \texttt{co-ord-last} is not ($0.027$ vs.\ $0.072$). Above
$\beta{=}3$, the \cers\ of these three sets falls and the random sets' spread widens, and at
$10^4$ the discovered sets lie within the random range, as in Table~\ref{tab:ncs-vs-cers}. With
5 random sets the rank-$p$ floor is $1/6$, so this pilot does not by itself establish
specificity at weak strength.

\begin{table}[h]
\centering
\scriptsize
\setlength{\tabcolsep}{2.5pt}
\caption{\cers\ against steering strength $\beta$ (Qwen3-VL-8B, flat top-100, $N{=}200$): the
discovered set $S$ and the mean [min, max] of 5 layer-matched random sets. $\beta{=}30$ and
$10^3$ lie between their neighbours and are omitted.}
\label{tab:beta}
\begin{tabular}{rcccccccc}
\toprule
& \multicolumn{2}{c}{\texttt{cm-ref-qsp}} & \multicolumn{2}{c}{\texttt{co-ref-qsp}} & \multicolumn{2}{c}{\texttt{cm-ord-last}} & \multicolumn{2}{c}{\texttt{co-ord-last}} \\
\cmidrule(lr){2-3}\cmidrule(lr){4-5}\cmidrule(lr){6-7}\cmidrule(lr){8-9}
$\beta$ & $S$ & random & $S$ & random & $S$ & random & $S$ & random \\
\midrule
1 & .279 & $-$.026 [$-$.071, .009] & .176 & .056 [$-$.021, .146] & .132 & $-$.040 [$-$.064, $-$.017] & .032 & .015 [$-$.013, .067] \\
3 & .869 & $-$.103 [$-$.279, .029] & .789 & .204 [$-$.059, .347] & .308 & $-$.134 [$-$.192, $-$.071] & .027 & .072 [$-$.076, .332] \\
10 & .317 & .004 [$-$.241, .221] & .396 & .270 [.087, .558] & .092 & $-$.068 [$-$.200, .088] & .182 & .175 [$-$.232, .461] \\
100 & .266 & $-$.001 [$-$.214, .218] & .273 & .226 [.081, .517] & .042 & $-$.040 [$-$.155, .095] & .206 & .173 [$-$.224, .421] \\
$10^4$ & .179 & .031 [$-$.291, .291] & .172 & .376 [.052, .841] & .018 & $-$.032 [$-$.192, .098] & .176 & .161 [$-$.202, .463] \\
\bottomrule
\end{tabular}
\end{table}

\subsection{Second-model necessity check}
\label{app:second-model}

We repeated discovery and the layer-matched NCS test for the four ordinal/referring $\times$
last-token/query-span settings on COCO strips with two further models ($N{=}400$ discovery
strips, $N{=}400$ evaluation items, top-100 heads, $R{=}20$).

\begin{table}[h]
\centering
\scriptsize
\setlength{\tabcolsep}{3pt}
\caption{Second-model check: NCS of the discovered head-set vs.\ the mean of 20 layer-matched
random controls (rank $p$), COCO discovery, top-100 heads, $N{=}400$. Bold: above all 20
controls.}
\label{tab:second-model}
\begin{tabular}{lccc}
\toprule
Setting (COCO) & Qwen3-VL-8B & Qwen3-VL-4B & Qwen2-VL-7B \\
\midrule
ordinal + last token & $-0.002$ vs.\ $0.000$ ($.86$) & $0.006$ vs.\ $0.003$ ($.14$) & $-0.002$ vs.\ $0.001$ ($1.00$) \\
ordinal + query span & $\mathbf{0.036}$ vs.\ $-0.003$ ($.048$) & $-0.005$ vs.\ $0.005$ ($.91$) & $0.001$ vs.\ $0.002$ ($.57$) \\
referring + last token & $-0.002$ vs.\ $0.002$ ($.76$) & $-0.004$ vs.\ $0.004$ ($.91$) & $\mathbf{0.024}$ vs.\ $0.001$ ($.048$) \\
referring + query span & $\mathbf{0.125}$ vs.\ $0.000$ ($.048$) & $\mathbf{0.054}$ vs.\ $0.002$ ($.048$) & $\mathbf{0.029}$ vs.\ $0.002$ ($.048$) \\
\bottomrule
\end{tabular}
\end{table}

On Qwen3-VL-4B the referring + query-span effect replicates at about $40\%$ of its Qwen3-VL-8B
size, above all 20 controls, and both last-token settings are null; the ordinal + query-span
effect does not replicate. On Qwen2-VL-7B the referring + query-span effect also replicates
($0.029$, about $23\%$ of its Qwen3-VL-8B size), both ordinal settings are null, and the
referring + last-token setting is weakly necessary ($0.024$, above all 20 controls).

\section{Discovery overlap}
\label{app:discovery}

\subsection{Noise ceiling}
\label{app:noise-ceiling}

To tell discovery differences from sampling noise, we recomputed each setting's score per
discovery strip (Qwen3-VL-8B, $N{=}400$ strips per corpus; recomputed top-100 sets match the
published ones on $98$--$100$ of $100$ heads) and drew $200$ random half-splits. Within a
setting, top-100 sets from the two halves share $96.2$--$99.3$ heads (the noise ceiling). Across
settings we compare halves of the same size ($200$ strips, disjoint halves when both settings
use the same corpus), so within- and cross-setting overlaps are directly comparable
(Table~\ref{tab:noise-ceiling}). Changing the corpus keeps a median of $83.8$ heads
($78.5$--$90.5$), changing the query type $54.6$ ($35.2$--$78.8$), and changing the attention
source $48.6$ ($35.8$--$79.1$); every cross-setting overlap lies well below its ceiling. The
ordinal + output-token settings are not included.

\begin{table}[h]
\centering
\scriptsize
\caption{Top-100 overlap between two discovery settings at $200$ strips per half (mean
[$2.5$, $97.5$ percentile] over $200$ half-splits), the geometric-mean noise ceiling of the two
settings, and the overlap as a fraction of that ceiling.}
\label{tab:noise-ceiling}
\begin{tabular}{llccc}
\toprule
What changes & Pair & Shared heads & Ceiling & Fraction \\
\midrule
corpus & \texttt{cm-ord-last} vs.\ \texttt{co-ord-last} & $81.2$ [$80$, $83$] & $98.4$ & $0.82$ \\
corpus & \texttt{cm-ord-qsp} vs.\ \texttt{co-ord-qsp} & $90.5$ [$90$, $91$] & $99.2$ & $0.91$ \\
corpus & \texttt{cm-ref-last} vs.\ \texttt{co-ref-last} & $83.8$ [$81$, $86$] & $96.6$ & $0.87$ \\
corpus & \texttt{cm-ref-out} vs.\ \texttt{co-ref-out} & $78.5$ [$76$, $80$] & $97.7$ & $0.80$ \\
corpus & \texttt{cm-ref-qsp} vs.\ \texttt{co-ref-qsp} & $83.8$ [$82$, $86$] & $97.8$ & $0.86$ \\
\midrule
query type & \texttt{cm-ord-last} vs.\ \texttt{cm-ref-last} & $35.2$ [$33$, $37$] & $97.2$ & $0.36$ \\
query type & \texttt{cm-ord-qsp} vs.\ \texttt{cm-ref-qsp} & $78.8$ [$78$, $80$] & $99.1$ & $0.79$ \\
query type & \texttt{co-ord-last} vs.\ \texttt{co-ref-last} & $36.5$ [$34$, $38$] & $97.8$ & $0.37$ \\
query type & \texttt{co-ord-qsp} vs.\ \texttt{co-ref-qsp} & $72.8$ [$71$, $75$] & $97.9$ & $0.74$ \\
\midrule
attention source & \texttt{cm-ord-last} vs.\ \texttt{cm-ord-qsp} & $39.6$ [$38$, $40$] & $98.7$ & $0.40$ \\
attention source & \texttt{cm-ref-last} vs.\ \texttt{cm-ref-qsp} & $45.5$ [$44$, $47$] & $97.6$ & $0.47$ \\
attention source & \texttt{cm-ref-last} vs.\ \texttt{cm-ref-out} & $50.6$ [$49$, $52$] & $97.1$ & $0.52$ \\
attention source & \texttt{cm-ref-qsp} vs.\ \texttt{cm-ref-out} & $79.1$ [$78$, $80$] & $98.6$ & $0.80$ \\
attention source & \texttt{co-ord-last} vs.\ \texttt{co-ord-qsp} & $35.8$ [$35$, $37$] & $98.9$ & $0.36$ \\
attention source & \texttt{co-ref-last} vs.\ \texttt{co-ref-qsp} & $46.6$ [$45$, $49$] & $96.7$ & $0.48$ \\
attention source & \texttt{co-ref-last} vs.\ \texttt{co-ref-out} & $51.4$ [$50$, $54$] & $97.1$ & $0.53$ \\
attention source & \texttt{co-ref-qsp} vs.\ \texttt{co-ref-out} & $69.8$ [$68$, $73$] & $96.9$ & $0.72$ \\
\bottomrule
\end{tabular}
\end{table}

\subsection{Crossover of query and attention source}
\label{app:d17}

Table~\ref{tab:crossover-main} holds corpus and layout fixed: $200$ strips of six COCO photos,
one photo per strip containing an object whose category is unique among the six panels, so a
referring query such as ``Find the truck.'' names exactly one panel. On this corpus we compute
all six query $\times$ source combinations with each published score's own code, changing only
which rows of the attention matrix are averaged (each panel's rectangle serves as the target
box). Each top-20 set is compared with the \emph{native} top-20 sets, discovered on each
method's own corpus. Table~\ref{tab:crossover-main} reports four combinations;
Table~\ref{tab:crossover-discovery} gives the remaining two. These overlaps use a single
discovery run; their split-half noise ceiling was not computed.

\begin{table}[h]
\centering
\small
\caption{Top-20 overlap with the native head-sets for the two crossover combinations not in
Table~\ref{tab:crossover-main} ($200$ COCO strips, Qwen3-VL-8B). Bold: the combination's own
native set.}
\label{tab:crossover-discovery}
\begin{tabular}{lccc}
\toprule
Query + attention source & ord-last & ref-qspan & ref-out \\
\midrule
ordinal + output tokens & $8$ & $8$ & $9$ \\
referring + output tokens (\virro) & $5$ & $7$ & $\mathbf{13}$ \\
\bottomrule
\end{tabular}
\end{table}

\subsection{Overlap between the published scores on other models}
\label{app:visual-verification}

\begin{figure}[h]
\centering
\includegraphics[width=0.98\linewidth]{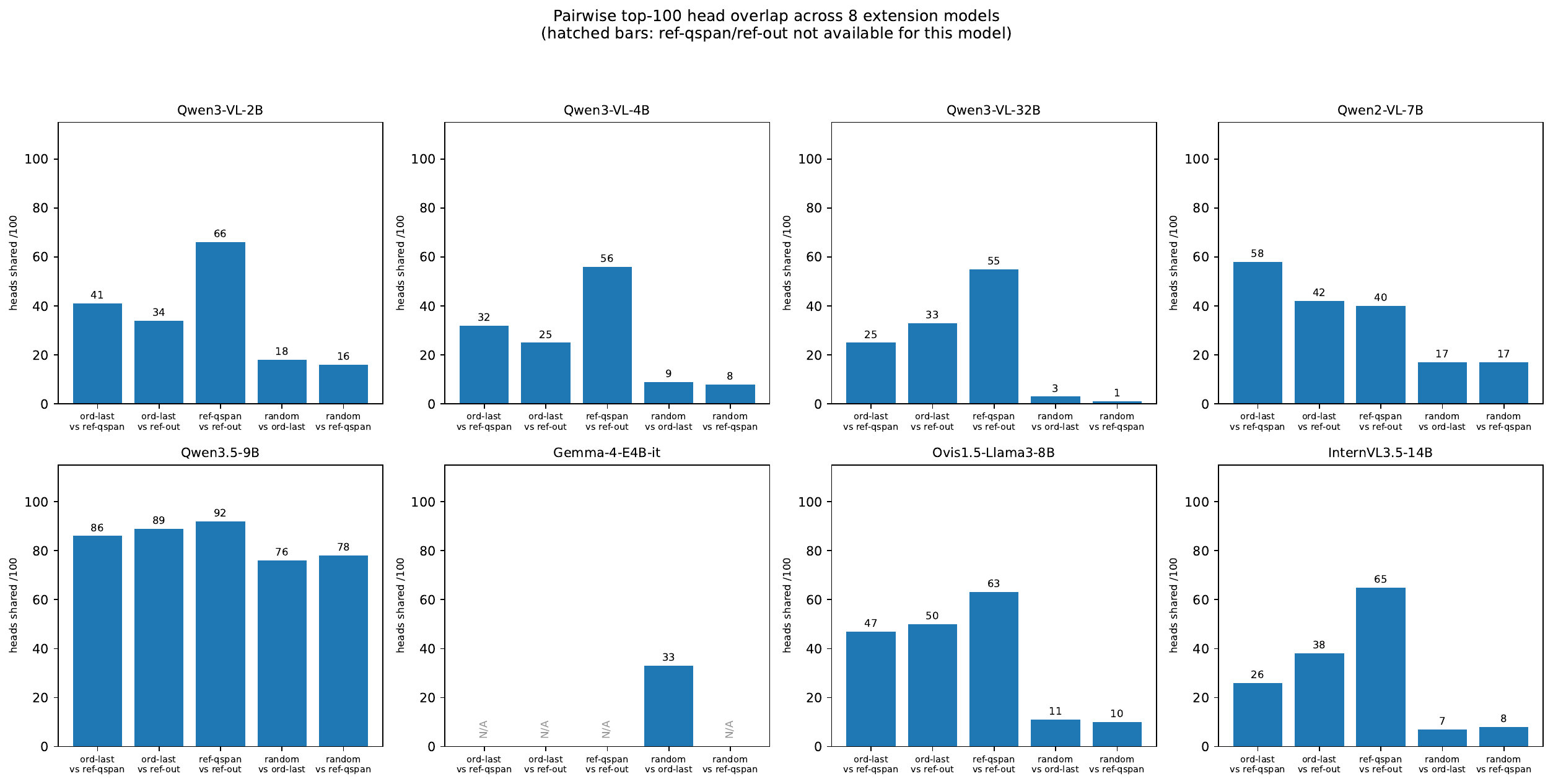}
\caption{Top-100 overlap between the native ord-last, ref-qspan and ref-out sets and a uniform
random draw of 100 heads, on eight further models (Gemma-4-E4B-it has ord-last scores only, so
its other bars are marked N/A). Ord-last/ref-qspan overlap is $25$--$86$/100 across the seven
scored models, ord-last/ref-out $25$--$89$/100, and ref-qspan/ref-out $40$--$92$/100; the random
draw shares $3$--$76$/100 with ord-last and $1$--$78$/100 with ref-qspan.}
\label{fig:overlap-allmodels}
\end{figure}

On Qwen3-VL-8B the native ord-last and ref-qspan sets share $2$ of their top-20 heads
(Table~\ref{tab:crossover-main}). Figure~\ref{fig:overlap-allmodels} repeats the comparison at
top-100 on eight further models, with each model's own discovery scores.

\section{Out-of-domain and cross-model evaluation}
\label{app:d20}

This section evaluates the head-sets of Table~\ref{tab:ncs-vs-cers} on benchmarks that share
no images or prompts with the discovery corpora. We use the four query-span head-sets (flat
top-100), each against a single layer-matched random draw, on eight datasets: MathVista,
CountBenchQA, CVBench, MMStar, SpatialRGPT, Spatial457, VAW-attribute, and ImageNet ($300$
distinct classes). \cers\ uses $2\times2$ composites of four independent real examples
(Appendix~\ref{app:pointing}), with every option a full (question, answer) pair; NCS uses the
matched/mismatched single-image design of \S\ref{sec:necessity-method}, with distractor options
of the correct answer's coarse type (boolean, numeric or free text). The NCS pair builder
targets $80$ pairs per dataset and skips a draw when too few type-matched distractors or no
valid mismatched image exist, giving $563$ pairs (MMStar 74, SpatialRGPT 54, Spatial457 35,
all others 80); no filtering on model correctness is applied.

\textbf{Baseline audit.} Before the head-set evaluation we ran an audit ($N{=}400$/dataset,
Table~\ref{tab:d20-audit}). For \cers, the oracle condition (grid cell named in the prompt)
should approach $1$ and the floor condition (blank image) should sit at $0.25$; for NCS the
mismatched arm should sit at $0.25$. SpatialRGPT's oracle is $0.28$ (its region markers do not
survive compositing), CVBench's and Spatial457's are $0.63$ and $0.57$, and the NCS mismatched
arm is above chance on MathVista ($0.73$), CVBench ($0.61$), Spatial457 ($0.53$), SpatialRGPT
($0.47$) and VAW ($0.47$). Causes include skewed answer vocabularies (CVBench's answer is ``1''
on about $80\%$ of items), boolean-heavy answer spaces (Spatial457's answer is ``False'' on
$53\%$ of items), and duplicate or near-synonymous options from the type-matched sampler.
Content-free calibration \citep{zhao2021calibrate} brings the mismatched arm near chance for
four of the five (not SpatialRGPT); we report the uncalibrated pipeline and compare every
discovered set with a random set scored on the same items.

\begin{table}[h]
\centering
\scriptsize
\caption{Out-of-domain baseline audit, Qwen3-VL-8B, $N{=}400$/dataset. Oracle: target cell named
in the prompt (expect $\approx1$). Floor: blank image (expect $0.25$). Matched/mismatched: NCS
arms without intervention (expect $\approx1$ / $0.25$). Bold: outside the expected range.}
\label{tab:d20-audit}
\begin{tabular}{lcccc}
\toprule
Dataset & \cers\ oracle & \cers\ floor & NCS matched & NCS mismatched \\
\midrule
MathVista & $0.885$ & $0.275$ & $0.870$ & $\mathbf{0.730}$ \\
CountBenchQA & $0.950$ & $0.302$ & $0.897$ & $0.220$ \\
CVBench & $\mathbf{0.632}$ & $0.268$ & $0.740$ & $\mathbf{0.605}$ \\
MMStar & $0.815$ & $0.273$ & $0.935$ & $0.367$ \\
SpatialRGPT & $\mathbf{0.282}$ & $0.230$ & $0.492$ & $\mathbf{0.470}$ \\
Spatial457 & $\mathbf{0.573}$ & $0.217$ & $0.735$ & $\mathbf{0.527}$ \\
VAW & $0.807$ & $0.253$ & $0.780$ & $\mathbf{0.468}$ \\
ImageNet & $0.993$ & $0.237$ & $1.000$ & $0.247$ \\
\bottomrule
\end{tabular}
\end{table}

\begin{table}[h]
\centering
\small
\caption{Out-of-domain \cers\ and NCS, pooled over the eight datasets, Qwen3-VL-8B ($N{=}1280$
per head-set for \cers, $563$ pairs for NCS; flat top-100). Each discovered set is followed by
its single layer-matched random draw. 95\% intervals: $t$ over items (\cers), bootstrap (NCS).}
\label{tab:d20-cers}
\begin{tabular}{lcc}
\toprule
Head-set & \cers\ [95\% CI] & NCS [95\% CI] \\
\midrule
cm-ref-qsp & $\mathbf{0.288}$ [$0.272$, $0.304$] & $\mathbf{0.225}$ [$0.188$, $0.264$] \\
\quad random & $-0.117$ [$-0.135$, $-0.100$] & $0.020$ [$0.001$, $0.039$] \\
co-ref-qsp & $0.092$ [$0.078$, $0.107$] & $0.211$ [$0.175$, $0.249$] \\
\quad random & $0.009$ [$-0.004$, $0.021$] & $0.023$ [$0.008$, $0.037$] \\
co-ord-qsp & $0.271$ [$0.257$, $0.286$] & $0.161$ [$0.127$, $0.196$] \\
\quad random & $-0.141$ [$-0.158$, $-0.124$] & $0.011$ [$-0.003$, $0.026$] \\
cm-ord-qsp & $0.216$ [$0.202$, $0.229$] & $0.143$ [$0.109$, $0.176$] \\
\quad random & $-0.100$ [$-0.115$, $-0.085$] & $0.017$ [$0.004$, $0.029$] \\
\bottomrule
\end{tabular}
\end{table}

Each discovered set's \cers\ and NCS interval lies above its random draw's. Three of the four
random draws have negative \cers\ ($-0.10$ to $-0.14$); relative to zero, the discovered sets'
\cers\ is $0.09$--$0.29$. Three of the four random draws have NCS slightly above zero
($0.017$--$0.023$). ImageNet, the dataset that passes the audit cleanly, shows the largest
discovered-vs.-random separation on \cers; on NCS the largest is CountBenchQA, which contributes
a large share of the pooled NCS difference.

\subsection{Five further models}
\label{app:crossmodel}

We repeat the out-of-domain evaluation on five further models with head-sets discovered anew
per model (flat top-100; the six settings of Table~\ref{tab:crossmodel-full}), each against one
layer-matched random draw. Models whose chat template reasons by default (Qwen3.5-9B) are run
with thinking disabled, so the forced-letter readout scores the answer letter rather than the
first token of a reasoning trace. InternVL3.5-14B is loaded in 4-bit (NF4).
Table~\ref{tab:crossmodel-main} summarizes this table.

\begin{table}[h]
\centering
\scriptsize
\setlength{\tabcolsep}{3pt}
\caption{Out-of-domain \cers\ / NCS of the discovered set, with its single layer-matched random
draw in parentheses, five further models (flat top-100, $N{=}1280$ \cers, $563$ NCS pairs).}
\label{tab:crossmodel-full}
\begin{tabular}{lccccc}
\toprule
Setting & Qwen3-VL-4B & Qwen3.5-9B & Gemma-4-E4B & Ovis1.5-8B & InternVL3.5-14B \\
\midrule
\multicolumn{6}{l}{\textit{\cers}} \\
cm-ref-qsp & $.098$ ($-.022$) & $-.003$ ($-.038$) & $.001$ ($.035$) & $.011$ ($-.026$) & $.246$ ($.135$) \\
co-ref-qsp & $.070$ ($-.041$) & $-.017$ ($-.041$) & $-.005$ ($.004$) & $.013$ ($-.003$) & $.341$ ($-.031$) \\
co-ord-qsp & $.154$ ($-.021$) & $-.017$ ($.040$) & $.000$ ($.038$) & $.005$ ($-.035$) & $.228$ ($.071$) \\
cm-ord-qsp & $.090$ ($-.041$) & $-.034$ ($.028$) & $.007$ ($-.026$) & $.014$ ($-.014$) & $.288$ ($.102$) \\
co-ref-out & $-.015$ ($-.020$) & $-.024$ ($.018$) & $-.002$ ($-.020$) & $.001$ ($-.001$) & $.379$ ($-.176$) \\
cm-ord-last & $.006$ ($-.053$) & $-.047$ ($.035$) & $-.010$ ($.075$) & $.049$ ($.088$) & $.205$ ($-.020$) \\
\midrule
\multicolumn{6}{l}{\textit{NCS}} \\
cm-ref-qsp & $.140$ ($.015$) & $.226$ ($.029$) & $.366$ ($.097$) & $.204$ ($.018$) & $.076$ ($.004$) \\
co-ref-qsp & $.211$ ($.021$) & $.217$ ($.025$) & $.378$ ($.076$) & $.254$ ($.023$) & $.081$ ($.033$) \\
co-ord-qsp & $.158$ ($.022$) & $.224$ ($.043$) & $.402$ ($.070$) & $.211$ ($.005$) & $.071$ ($.023$) \\
cm-ord-qsp & $.161$ ($.032$) & $.218$ ($.030$) & $.401$ ($.071$) & $.220$ ($.007$) & $.084$ ($.018$) \\
co-ref-out & $.159$ ($.006$) & $.231$ ($.043$) & $.388$ ($.081$) & $.236$ ($.012$) & $.133$ ($.015$) \\
cm-ord-last & $.010$ ($.002$) & $.223$ ($.060$) & $.022$ ($.035$) & $.024$ ($.004$) & $.024$ ($.012$) \\
\bottomrule
\end{tabular}
\end{table}

All 25 query-span and output-token sets exceed their random draw on NCS. Steering is
model-dependent: \cers\ is within $\pm0.05$ of zero for every set on Qwen3.5-9B, Gemma-4-E4B and
Ovis1.5-8B, while InternVL3.5-14B shows large values ($0.21$--$0.38$) with single random draws
up to $0.135$.

\section{The 14-task pointing suite}
\label{app:pointing}

\textbf{Construction.} The suite converts seven VQA benchmarks (CV-Bench, MMStar, MathVista,
CountBenchQA, SpatialRGPT-Bench, Spatial457, VAW attributes) into the $2\times2$ \cers\ format in
two ways (Table~\ref{tab:tasksexplained}). The seven \emph{converted} tasks build each grid from
one benchmark item by image retrieval or same-image overlay (Appendix~\ref{app:datasets}). The
seven \emph{composite} tasks place four independent, already-solved real examples (own image, own
question, own ground truth) in one grid; steering toward cell $T$ is scored by the forced
four-way probability of $T$'s own answer, with the four cells' own answers as options, so no
distractor is constructed. Three converted tasks were flagged before the comparison for
construction shortcuts introduced by similar-looking distractors (spatial457\_pointing,
MathVista\_pointing, attribute\_identification). On MathVista the model also answers largely
from text: accuracy on the composite MathVista task is $0.850$ with a blank gray image and
$0.825$ with the real image ($N{=}40$).

\begin{table}[h]
\centering
\scriptsize
\caption{The 14 tasks of the pointing suite. Every task uses the same $2\times2$ \cers\
apparatus (Eq.~\ref{eq:cers}); only how the four cells and four options are built differs.}
\label{tab:tasksexplained}
\begin{tabular}{@{}p{2.6cm}p{1.5cm}p{8.7cm}@{}}
\toprule
Task & Family & What the model is shown and asked \\
\midrule
CV-Bench\_pointing & converted & Real photo; question about counting or relations; wrong answers overlaid on three copies of the same photo. \\
MMStar\_pointing & converted & Same recipe, with MMStar's multi-domain VQA items. \\
SpatialRGPT\_pointing & converted & Real photo with two labeled regions and a spatial-relation question; distractor answers overlaid on copies. \\
Spatial457\_pointing & converted & Four synthetic scene renders; the question refers to one scene's objects; the correct cell is that scene. \\
CountBenchQA\_pointing & converted & Real photo of counted objects; ``how many X''; candidate counts as text options. \\
MathVista\_pointing & converted & Four math diagrams; the correct item's question and choices asked in text. \\
attribute\_identification & converted & Real VAW photo; the object's box is the steering target; candidate attribute words as options. \\
general\_causal\_* (7 tasks) & composite & Four independent, already-solved real examples from one source dataset composited $2\times2$; the target cell's own question is asked, with the four cells' own answers as options. \\
\bottomrule
\end{tabular}
\end{table}

\textbf{Discovered vs.\ random.} Table~\ref{tab:generalcausal} compares the flat top-20 \vir\
set with one layer-matched random draw ($N{=}80$ per task). Pooled over all 14 tasks \vir\
scores $0.142$ ($[0.129,0.156]$) and the random set $0.115$ ($[0.101,0.130]$); the paired
difference is $0.027$ ($[0.016,0.038]$, $p=2.1\times10^{-6}$). Without the three flagged tasks
\vir\ scores $0.139$ ($[0.125,0.154]$) and the random set $0.102$ ($[0.087,0.117]$), $73\%$ of
\vir's value; the paired difference is $0.037$ ($[0.025,0.049]$, $p=3.2\times10^{-10}$,
$N{=}880$). On the same 11 tasks the flat top-20 \virrq\ set scores $0.081$. The random set
matches or exceeds \vir\ on spatial457\_pointing and MathVista\_pointing and ties it on
general\_causal\_MathVista.

\begin{table}[h]
\centering
\footnotesize
\caption{\cers\ of the flat top-20 \vir\ set and one layer-matched random draw, Qwen3-VL-8B,
$N{=}80$ per task. Bold: the three flagged tasks. The gap is computed from unrounded values.}
\label{tab:generalcausal}
\begin{tabular}{lccc}
\toprule
Task & \vir\ & Random & Gap \\
\midrule
\textbf{spatial457\_pointing} & $0.427$ & $0.459$ & $-0.033$ \\
\textbf{MathVista\_pointing} & $0.012$ & $0.028$ & $-0.016$ \\
\textbf{attribute\_identification} & $0.017$ & $0.004$ & $0.012$ \\
\midrule
general\_causal\_MathVista & $0.069$ & $0.069$ & $0.000$ \\
CountBenchQA\_pointing & $0.019$ & $0.013$ & $0.006$ \\
general\_causal\_CVBench & $0.197$ & $0.186$ & $0.011$ \\
general\_causal\_MMStar & $0.208$ & $0.196$ & $0.011$ \\
MMStar\_pointing & $0.229$ & $0.196$ & $0.034$ \\
SpatialRGPT\_pointing & $0.177$ & $0.133$ & $0.043$ \\
general\_causal\_VAW & $0.104$ & $0.057$ & $0.047$ \\
general\_causal\_SpatialRGPT & $0.125$ & $0.074$ & $0.052$ \\
general\_causal\_Spatial457 & $0.128$ & $0.074$ & $0.054$ \\
CV-Bench\_pointing & $0.112$ & $0.054$ & $0.058$ \\
general\_causal\_CountBenchQA & $0.164$ & $0.071$ & $0.093$ \\
\midrule
All 14 tasks & $0.142$ & $0.115$ & $0.027$ \\
11 unflagged tasks & $0.139$ & $0.102$ & $0.037$ \\
\bottomrule
\end{tabular}
\end{table}

\subsection{Other models}
\label{app:d11}

We ran the same \cers\ measurement on eight further models with the native \vir, \virrq\ and
\virro\ scores (flat top-20; one layer-matched random draw matched to \vir's per-layer counts;
all 14 tasks, $N{=}1120$ per model; Table~\ref{tab:necsuff-allmodels}). Qwen3-VL-32B and
InternVL3.5-14B were loaded in 4-bit (NF4), and Qwen3.5-9B was run with thinking disabled.
\vir's \cers\ is above zero on seven of the eight models (not Ovis1.5-Llama3-8B), and the random
draw's \cers\ is negative or at most $0.003$ on all eight. The margin over the random draw is
clear on Qwen3-VL-2B/4B/32B, InternVL3.5-14B and Qwen3.5-9B, modest on Qwen2-VL-7B, and
negligible on InternVL3.5-8B ($0.006$ vs.\ $0.003$).

\begin{table}[h]
\centering
\scriptsize
\caption{Pointing-suite \cers\ on eight further models (all 14 tasks, $N{=}1120$ per model, flat
top-20, one random draw per model). Bold: $p<0.05$ against zero (one-sample $t$-test).}
\label{tab:necsuff-allmodels}
\begin{tabular}{lccccc}
\toprule
Model & Baseline acc.\ & \vir\ & \virrq\ & \virro\ & Random \\
\midrule
Qwen3-VL-2B & $0.492$ & $\mathbf{0.161}$ & $\mathbf{0.033}$ & $\mathbf{0.022}$ & $\mathbf{-0.041}$ \\
Qwen3-VL-4B & $0.521$ & $\mathbf{0.127}$ & $\mathbf{0.189}$ & $\mathbf{0.066}$ & $\mathbf{-0.010}$ \\
Qwen2-VL-7B & $0.487$ & $\mathbf{0.066}$ & $\mathbf{0.027}$ & $\mathbf{0.025}$ & $\mathbf{-0.056}$ \\
Qwen3-VL-32B (4-bit) & $0.537$ & $\mathbf{0.122}$ & $\mathbf{0.168}$ & $\mathbf{0.116}$ & $\mathbf{-0.020}$ \\
InternVL3.5-8B & $0.527$ & $\mathbf{0.006}$ & $\mathbf{0.019}$ & $\mathbf{0.075}$ & $\mathbf{0.003}$ \\
InternVL3.5-14B (4-bit) & $0.483$ & $\mathbf{0.102}$ & $\mathbf{0.150}$ & $\mathbf{0.159}$ & $\mathbf{-0.054}$ \\
Ovis1.5-Llama3-8B & $0.478$ & $-0.001$ & $\mathbf{0.032}$ & $\mathbf{0.031}$ & $\mathbf{-0.030}$ \\
Qwen3.5-9B & $0.557$ & $\mathbf{0.058}$ & $\mathbf{0.070}$ & $\mathbf{0.139}$ & $\mathbf{-0.080}$ \\
\bottomrule
\end{tabular}
\end{table}

\section{Method details}
\label{app:method-details}

\subsection{CERS and NCS illustrated}
\label{app:cers-ncs-diagram}

\begin{figure}[h]
\centering
\includegraphics[width=\linewidth]{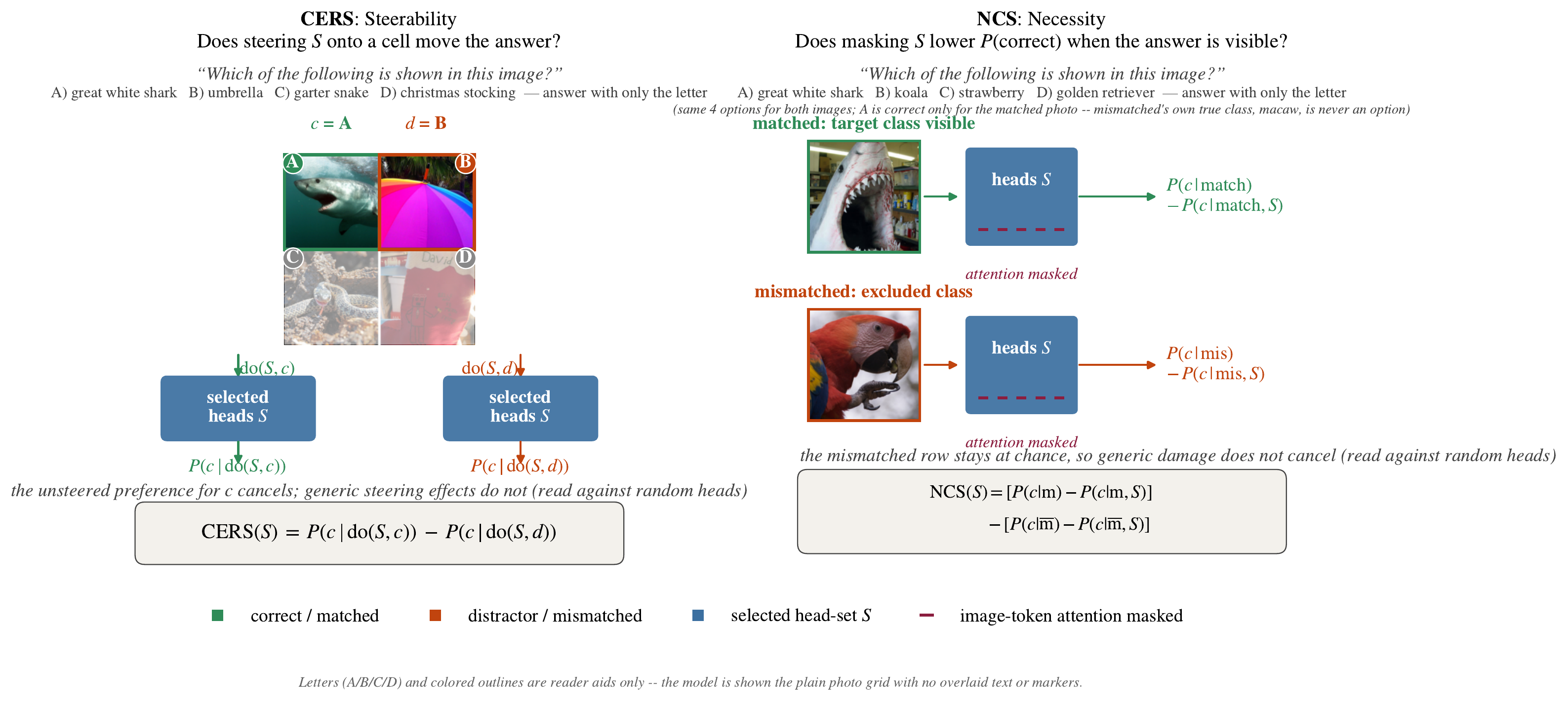}
\caption{\cers\ vs.\ NCS on real example images. Left: the $2\times2$ MCQ grid with the correct
cell (A, green) and a distractor cell (B, orange), steered via $\mathrm{do}(S,c)$ and
$\mathrm{do}(S,d)$. Right: the matched (target class visible, green) and mismatched (excluded
class, orange) images used for NCS, with the attention mask indicated below each head-set box.}
\label{fig:cers-ncs-diagram}
\end{figure}

Figure~\ref{fig:cers-ncs-diagram} illustrates both tests on real images. \cers\ (left) steers
the same $S$ once toward the correct cell and once toward a distractor cell (Eq.~\ref{eq:cers})
and subtracts the two forced-choice probabilities. NCS (right) applies the same whole-image
attention mask to a matched image (answer visible) and a mismatched image (answer absent) and
subtracts the two probability drops (\S\ref{sec:necessity-method}). Masking adds $-10^4$ to the pre-softmax
attention logits of the heads in $S$ at the image positions; every other head and the image
tokens themselves are unchanged.

\subsection{Per-layer head selection}
\label{app:algorithm}

Flat top-$k$ selection ranks all heads of $M$ jointly, so a few layers can dominate. The
per-layer alternative used in Table~\ref{tab:selection-position} (right) takes a fixed number of
heads from every layer (Algorithm~\ref{alg:select}); its random controls keep the same per-layer
counts.

\begin{algorithm}[h]
\caption{Per-layer head selection with layer-matched random controls}
\label{alg:select}
\begin{algorithmic}[1]
\STATE \textbf{Input:} score matrix $M \in \mathbb{R}^{L\times H}$, per-layer quota
$k_{\text{per\_layer}}$, number of controls $R$
\FOR{each layer $\ell = 1, \dots, L$}
  \STATE add the $k_{\text{per\_layer}}$ heads with the highest $M_{\ell,h}$ to $S$
\ENDFOR
\FOR{$r = 1, \dots, R$}
  \STATE for each layer, draw as many heads as $S$ holds there, uniformly from that layer's
  heads not in $S$; call the result $S_r$
\ENDFOR
\STATE \textbf{Return:} $S$, $\{S_r\}_{r=1}^R$
\end{algorithmic}
\end{algorithm}

\subsection{Evaluation protocol for every reported result}
\label{app:protocol}

Table~\ref{tab:protocol} lists the settings behind each result. All interventions use
$\beta=10^4$ (Eq.~\ref{eq:steer}) except in the strength pilot; masking means $-\beta$ on the image tokens for the heads in
$S$ in a single forward pass; all readouts are forced-choice letter probabilities.

\begin{table}[h]
\centering
\scriptsize
\caption{Protocol per result, Qwen3-VL-8B-Instruct unless stated. $R$: random head-sets per
discovered set.}
\label{tab:protocol}
\begin{tabular}{@{}p{3.6cm}p{2.4cm}p{4.1cm}p{1.1cm}p{0.9cm}@{}}
\toprule
Result & Head-set & Evaluation data & $N$ & $R$ \\
\midrule
Tables~\ref{tab:noise-main}, \ref{tab:noise-ceiling} (noise ceiling) & flat top-100 & discovery strips, 200 half-splits & 400 & -- \\
Tables~\ref{tab:crossover-main}, \ref{tab:crossover-discovery}; Fig.~\ref{fig:overlap-allmodels} & top-20; top-100 & 200 COCO strips (discovery only) & 200 & -- \\
Table~\ref{tab:ncs-vs-cers}, Fig.~\ref{fig:ncs-vs-cers}, Tables~\ref{tab:e1-random-control}, \ref{tab:ncs-arm-audit} & flat top-100 & ImageNet $2\times2$ grid (\cers); ImageNet MCQ (NCS) & 400 & 20 (250) \\
Table~\ref{tab:necessity-checks}, \ref{tab:ksweep}, Fig.~\ref{fig:ksweep-main} & flat top-$k$, $k\in\{20..160\}$ & ImageNet MCQ & 400 & 20 (250) \\
Table~\ref{tab:e2-stress} & 6 head-sets & ImageNet MCQ + text-solvable arm & 400 & -- \\
Tables~\ref{tab:selection-position}, \ref{tab:position-full} & flat top-100; $k_{\mathrm{per\_layer}}{=}3$ & ImageNet MCQ (\cers: grid) & 400 & 20 \\
Table~\ref{tab:beta} (pilot) & flat top-100, $\beta\in[1,10^4]$ & ImageNet $2\times2$ grid & 200 & 5 \\
Table~\ref{tab:shared-unique} & shared / specific parts of top-100 & ImageNet MCQ & 400 & 20 \\
Table~\ref{tab:clean-grid-main}, \ref{tab:clean-grid}; Fig.~\ref{fig:mass-scores} & flat top-100 & leak-audited ImageNet grid & 120 & 100 \\
Table~\ref{tab:second-model} & flat top-100 & ImageNet MCQ & 400 & 20 \\
Tables~\ref{tab:ood-main}, \ref{tab:d20-cers}, \ref{tab:crossmodel-main}, \ref{tab:crossmodel-full} & flat top-100 & 8 benchmarks & 1280 / 563 & 1 \\
Tables~\ref{tab:generalcausal}, \ref{tab:necsuff-allmodels} & flat top-20 & 14-task pointing suite & 1120 & 1 \\
\bottomrule
\end{tabular}
\end{table}

\subsection{Benchmark conversion}
\label{app:datasets}

Table~\ref{tab:dataset-detail} gives the recipe used for each converted benchmark.

\begin{table}[h]
\centering
\small
\caption{Source and conversion recipe of the seven converted benchmarks.}
\label{tab:dataset-detail}
\begin{tabular}{@{}p{3.2cm}p{4.0cm}p{5.0cm}@{}}
\toprule
Benchmark & Source content & Recipe \\
\midrule
CountBenchQA & counting VQA & same-image overlay (near-miss count distractors) \\
CV-Bench & single-image MCQ perception & same-image overlay \\
MathVista & MCQ diagram reasoning & same-image overlay (overlaid answer-choice text) \\
MMStar & single-image MCQ perception & same-image overlay \\
Spatial457 & spatial-relation VQA & image retrieval \\
SpatialRGPT-Bench & spatial-relation VQA & image retrieval \\
Attribute identification & VAW attributes + Visual Genome images & same-image overlay (cropped instance, overlaid candidate attribute) \\
\bottomrule
\end{tabular}
\end{table}

Two recipes are used, chosen per benchmark by the leak audit
\textbf{Image retrieval}: the distractor cells show different
photographs, domain-matched where a domain key is available; used where the baseline without the
target cell is near chance. \textbf{Same-image overlay}: all cells share one base image and differ
only in overlaid answer text; used for benchmarks that stay above chance under image retrieval,
where distractor images would reveal the answer through style rather than content.

\end{document}